\documentclass{article} %
\usepackage{iclr2027_conference,times}

\usepackage{hyperref}
\usepackage{url}

\title{ACID: Action Consistency via Inverse Dynamics for Planning with World Models}

\author{
Gawon Seo$^1$\quad
Dongwon Kim$^2$\setcounter{footnote}{1}\thanks{Corresponding authors.}\quad
Suha Kwak$^1$\footnotemark[\value{footnote}]\\[4pt]
$^1$POSTECH\quad
$^2$KAIST\\[2pt]
{\tt\small \url{https://gawon1224.github.io/ACID/}}
}

\usepackage{xcolor}
\usepackage{soul}

\newcommand{\modelname}{ACID\xspace}
\newcommand{\algnew}[1]{\textcolor{blue}{#1}} %

\usepackage{duckuments}
\usepackage{url}            %
\usepackage{booktabs}       %
\usepackage{amsfonts}       %
\usepackage{nicefrac}       %
\usepackage{microtype}      %
\usepackage{xcolor}         %
\usepackage{bbding}
\usepackage{multirow}
\usepackage{algorithm}
\usepackage{algpseudocode}
\usepackage{graphicx, amsmath, amssymb, caption, subcaption, overpic, textpos}
\usepackage{arydshln} %
\usepackage{listings}
\usepackage{setspace}
\usepackage{xspace}
\usepackage[table]{xcolor}
\usepackage{amsthm}
\usepackage{enumitem}
\usepackage{makecell}
\usepackage{wrapfig}
\usepackage{tikz}

\newtheorem{definition}{Definition}
\newtheorem{proposition}{Proposition}
\newtheorem{lemma}{Lemma}

\newtheorem{assumption}{Assumption}

\usepackage{amsmath,amsfonts,bm}

\def\Figref#1{Fig.~\ref{#1}}

\def\secref#1{Sec.~\ref{#1}}
\def\Secref#1{Sec.~\ref{#1}}

\def\eqref#1{equation~\ref{#1}}
\def\Eqref#1{Equation~\ref{#1}}

\def\1{\bm{1}}

\DeclareMathAlphabet{\mathsfit}{\encodingdefault}{\sfdefault}{m}{sl}
\SetMathAlphabet{\mathsfit}{bold}{\encodingdefault}{\sfdefault}{bx}{n}

\newcommand{\sparallel}{\smash[t]{\parallel}}

\iclrfinalcopy %

\begin{document}
\maketitle
\lhead{Preprint.}  %

\begin{abstract}
Decision-time planning with action-conditioned world models has become a popular paradigm for embodied control. However, the standard planning cost judges a candidate solely by how close its predicted terminal state lies to the goal, leaving the realizability of the intermediate transitions unchecked---a predicted trajectory can look convincing while the environment rollout drifts away from it. In this paper, we propose \modelname{}, a decision-time planning framework that introduces \emph{cycle action consistency}: the action inferred backward from a predicted transition by an inverse dynamics model should recover the one that was conditioned on. We fold this per-step residual into the planning cost via a scale-invariant adaptive weight. 
Across four action-conditioned world models and eight tasks 
encompassing object manipulation and articulated control in simulation, visual navigation, and real-robot manipulation, \modelname{} consistently improves planning and matches the baseline's accuracy with substantially less planning compute.
\end{abstract}

\section{Introduction}
\label{sec:intro}

World models~\citep{ha2018world} have become central to embodied decision making, serving as learned simulators that capture environmental dynamics within internal representations. Among the various ways 
of exploiting world models,
decision-time planning through test-time optimization has emerged as a particularly attractive paradigm since it requires no additional policy learning and accommodates a wide variety of world model designs~\citep{bar2025navigation,kim2026planning, zhou2024dino, maes2026leworldmodel, assran2025v, sobal2026learning, nam2026causal}. 
A popular instantiation of this paradigm is search-based planning with an action-conditioned world model~\citep{bar2025navigation, zhou2024dino, hansen2023td}: candidate action sequences are sampled and simulated through the world model, and the trajectory that minimizes a planning cost is executed in the environment.
The model predictive control (MPC)~\citep{Morari1999ModelPC} framework with cross-entropy method (CEM)~\citep{de2005cem} optimization follows precisely this procedure and has become a standard choice for decision-time planning with action-conditioned world models.

Despite its appeal, decision-time planning with world models 
{relies blindly on}
predicted trajectories without ever checking that they are \textit{realizable},
{that is, whether each transition could actually be produced by the conditioning action}
in the environment. 
A conventional planning cost defined solely by the proximity between the goal and the terminal state cannot see this: realizability is a property of the transition, and visual plausibility at the end does not imply it.
In other words, a prediction can look convincing while the actual environment rollout deviates from it.
Whether the intermediate transitions are realizable, or the trajectory merely drifts toward a goal-like state, is never considered by the cost. 
A planner can thus commit to action sequences whose predicted goal-reaching cannot be reproduced in the environment.

Recent work has sought to close this realization gap from the world model side, \emph{e.g.}, scaling up to large video-generative backbones~\citep{bar2025navigation, kim2026planning, chen2024diffusion, du2023video, hu2023gaia, gao2024vista, ye2026world, bruce2024genie, du2023learning} or enforcing action conditioning through guidance~\citep{kim2025freeaction, song2025history}.
These approaches either introduce prohibitive training cost or require a specific probabilistic formulation of the world model; the latter does not apply to the recent JEPA line of work, whose predictors model latent transitions deterministically~\citep{zhou2024dino, maes2026leworldmodel, assran2025v,  sobal2026learning, nam2026causal}.
Moreover, none addresses the planning cost: regardless of how faithful the predicted trajectories become, the objective still scores them solely by the final state proximity and cannot tell which ones genuinely realize their conditioning actions. What is needed is a complementary, decision-time mechanism that verifies action fidelity directly within the planning cost, while leaving the world model untouched and therefore composable with any of the above improvements. This raises a central question: without retraining the world model, how can the planning cost tell a trajectory that actually reaches the goal from one that only appears to?

Telling the two apart requires estimating the world model's rollout error at planning time, which cannot be done in the state space: the true outcome of a candidate action is unknown until executed. 
The action space, in contrast, has a known reference: the conditioning action itself. 
An inverse dynamics model (IDM) learns the same environment dynamics from the same transitions but reads them out in the action space, 
inferring the action that causes the transition between two consecutive states.
Chaining the world model and the IDM forms a cycle: the world model rolls a conditioning action forward into a predicted next state, and the IDM maps the resulting transition back to an action.
If the predicted transition is realizable, the inferred action recovers the conditioning action and the cycle closes; otherwise, the rollout error surfaces as a residual between the two actions.
We call this agreement \textit{cycle action consistency}, which turns a state-space error that cannot be measured into an action-space residual that can.
Building on this principle, we introduce \textbf{A}ction \textbf{C}onsistency via \textbf{I}nverse \textbf{D}ynamics (\textbf{ACID}), a decision-time planning framework that
introduces cycle action consistency as another planning cost,
preferring candidate action sequences whose predicted trajectories not only reach the goal but also remain realizable at every step.
In doing so, we repurpose the IDM as a decision-time verifier, estimating the world model's rollout error, rather than the offline action decoder or pseudo-labeler it has conventionally been.

By augmenting decision-time planning with cycle action consistency, \modelname delivers consistent gains across four action-conditioned world models, spanning three JEPA-style latent predictors and one video-generative model, and 
eight tasks encompassing object manipulation and articulated control in simulation, visual navigation, and tabletop manipulation with real robots.
These gains are insensitive to hyperparameter settings and come with substantially less total planning compute than the baseline.

\section{Related Work}
\label{sec:related}

\subsection{World models}

World models that learn environment dynamics and predict future states have rapidly diversified.
We group prior work into three families by how each obtains executable actions.
Action-conditioned world models take a current state and action as input and predict the resulting next state, and therefore rely on test-time search or optimization to find goal-reaching action sequences~\citep{maes2026leworldmodel, assran2025v, nam2026causal, bar2025navigation, rigter2024avid}.
They span latent predictors such as JEPA-based models as well as large video backbones.
State-only generative models take no action as input, but instead synthesize a goal- or language-conditioned sequence of future states and recover the action at each timestep afterward~\citep{du2023learning, xie2025latent}.
World action models instead generate future states and their corresponding actions together over a horizon, so a single forward pass emits an action sequence already aligned with the predicted visual future~\citep{ye2026world, ye2026gigaworld, zhu2025unified}.
This removes the need for both a separate inverse dynamics model and test-time search.
Our work targets action-conditioned world models, where the action is left to test-time search and a planning cost can intervene; \modelname{} adds its consistency cost at this decision-time stage without altering the model.

\subsection{Learning inverse dynamics for robot learning}
The inverse dynamics model (IDM) maps a pair of consecutive states to the action that produced the transition between them, and is widely used in robot learning.
In prior work, the IDM has typically been cast in three roles, none of which influences which action sequence the planner selects: they operate after the planner has already chosen a trajectory, or only during training.
As an action decoder, an IDM extracts executable actions from the output of a state-only or video-only planner.
UniPi~\citep{du2023learning} first generates a future video and LDP~\citep{xie2025latent} first generates a latent state trajectory, after which an IDM recovers the actions; the trajectory is already chosen, and the IDM only decodes it.
As an auxiliary task, inverse-dynamics prediction supplies a loss term that shapes representations during training~\citep{brandfonbrener2023inverse}, with no role at inference.
As a pseudo-labeler, an IDM assigns action labels to otherwise action-free video, enabling policy pretraining~\citep{baker2022video,jang2025dreamgen}.
In contrast, we cast the IDM as a \emph{verifier} at decision time, where its consistency signal enters the planning cost directly and shapes which action sequence the planner commits to.

\begin{figure}[t!]
    \centering
    \includegraphics[width=1.00\linewidth]{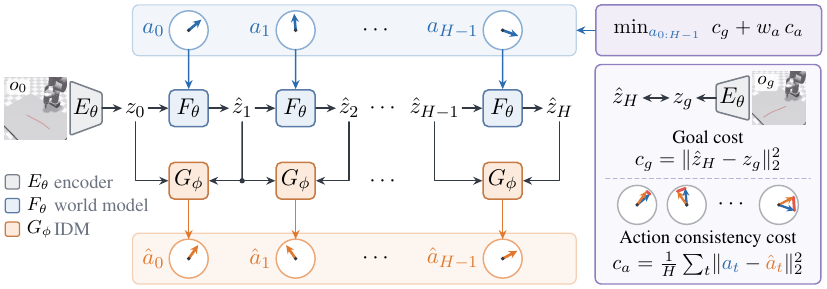}
    \caption{\textbf{Overall pipeline of \modelname.}
    (\textit{Left}): The encoder $E_\theta$ maps $o_0$ to $z_0$, and the world model $F_\theta$ rolls out $\hat{z}_{1:H}$ under a candidate action sequence $a_{0:H-1}$, colored in blue. The IDM $G_\phi$ infers from each predicted transition $(\hat{z}_t, \hat{z}_{t+1})$ the action $\hat{a}_t$, colored in orange, that explains the transition.
    (\textit{Right}): CEM searches for the action sequence minimizing $c_g + w_a c_a$, where the goal cost $c_g$ compares $\hat{z}_H$ with the encoded goal $z_g$, and the action consistency cost $c_a$ averages the per-step error between $a_t$ and $\hat{a}_t$.
    }
    \vspace{-5mm}
    \label{fig:overall}
\end{figure}
\section{Method}
\modelname augments decision-time planning with a per-step measure of how realizable a world model's predicted transitions are, as illustrated in Fig.~\ref{fig:overall}. We first formulate world models and MPC with CEM-based planning (\Secref{sec:preliminary}), introduce cycle action consistency and the IDM that verifies it (\Secref{sec:act_con}), and fold this signal into the planning cost with a scale-invariant adaptive weight (\Secref{sec:planning}).

\subsection{Preliminaries}
\label{sec:preliminary}

\textbf{Action-conditioned world model.}
An action-conditioned world model captures the environment's transition dynamics: given a current observation and an action, it predicts the resulting next observation. By chaining such one-step predictions, the model can forecast the outcome of an entire action sequence, enabling multi-step prediction.
Rather than modeling these dynamics in pixel space, we primarily adopt the JEPA-style design of recent latent world models~\citep{zhou2024dino, maes2026leworldmodel,sobal2026learning}, which learn transitions in a compact latent space. 
Concretely, an encoder $E_\theta$ maps a raw image observation $o_t$ to a latent state $z_t = E_\theta(o_t)$, and a world model $F_\theta$ predicts the next latent state $\hat{z}_{t+1}$ from $z_t$ given action $a_t$, \emph{i.e.}, $\hat{z}_{t+1} = F_\theta(z_t, a_t)$. 
The same formulation also applies to pixel-space models by taking $E_\theta$ to be the identity, which we revisit in our experiments.
{$F_\theta$ is trained with trajectories collected offline in advance}
without reward or task labels.
Our work targets
goal-conditioned visual control, where
the agent must reach a goal specified by an RGB image $o_g$. We address this task by planning action sequences through the learned world model {$F_\theta$}.

\textbf{Decision-time planning with action-conditioned world models.}
We instantiate decision-time planning as MPC over action sequences optimized with CEM. 
Given a current observation $o_0$ and a goal observation $o_g$, both are encoded into latent states $z_0 = E_\theta(o_0)$ and $z_g = E_\theta(o_g)$. For a candidate action sequence $a_{0:H-1}$ over a planning horizon of $H$ steps, 
the world model autoregressively predicts a latent trajectory:
\begin{equation}
\hat{z}_{t+1} = F_\theta(\hat{z}_t, a_t), \qquad \hat{z}_0 = z_0.
\end{equation}
Each candidate is then scored by the goal cost $c_g$ measuring its proximity to the goal,
\begin{equation}
\label{eq:latent-cost}
c_g(a_{0:H-1}) \;=\; \big\lVert \hat{z}_H - z_g \big\rVert_2^2.
\end{equation}
CEM optimizes this cost by iteratively sampling action sequences from a Gaussian distribution and {in turn fitting the distribution to} the top-$K$ {elite} candidates under $c_g$.
Following MPC, the optimized $H$-step action sequence is executed in the environment, after which the next $H$-step action sequence is replanned using the resulting observation.

\subsection{Cycle action consistency via inverse dynamics}
\label{sec:act_con}

\textbf{Cycle action consistency.}
We introduce \emph{cycle action consistency}, a verifier-based cost that detects when a predicted trajectory drifts from its conditioning actions: if $\hat{z}_{t+1} = F_\theta(\hat{z}_t, a_t)$ truly reflects $a_t$, then the action inferred from the transition $(\hat{z}_t, \hat{z}_{t+1})$ should recover $a_t$. As an external verifier that performs this backward inference, we introduce an
IDM, denoted by $G_\phi$, that infers the action responsible for a latent transition, $\hat{a}_t = G_\phi({z}_t, {z}_{t+1})$.
At each step of the predicted trajectory, we measure the \emph{per-step consistency residual} between the conditioning action and the action recovered by the verifier: $\big\lVert\, a_t \;-\; G_\phi\big(\hat{z}_t,\, \hat{z}_{t+1}\big) \,\big\rVert_2^2$.
This residual vanishes when the forward prediction $\hat{z}_{t+1} = F_\theta(\hat{z}_t, a_t)$ is the transition $a_t$ actually produces, and 
grows as the predicted transition drifts from the conditioning action.
The action consistency cost of a candidate sequence $a_{0:H-1}$ aggregates this residual over the planning horizon,
\begin{equation}
\label{eq:cyc-cost}
c_a(a_{0:H-1}) \;=\; \frac{1}{H} \sum_{t=0}^{H-1} \big\lVert\, a_t \;-\; G_\phi\big(\hat{z}_t,\, \hat{z}_{t+1}\big) \,\big\rVert_2^2.
\end{equation}
This cost computation does not require additional rollout in the world model since it reuses the predicted trajectory that was used to evaluate $c_g$ in Eq.~(\ref{eq:latent-cost}).

\textbf{Why the residual tracks realizability.}
A predicted step is \emph{realizable} if executing $a_t$ from $\hat{z}_t$ actually produces $\hat{z}_{t+1}$. We measure how far it is from being realizable by the \emph{realizability error} $e_t = \hat{z}_{t+1} - f(\hat{z}_t, a_t)$, where $f$ denotes the true latent dynamics, so that the step is realizable if and only if $e_t = 0$.
To relate $e_t$ to the per-step consistency residual, we assume locally linear latent dynamics $f(z, a) = Az + Ba$~\citep{wang2026temporal} in which distinct actions produce distinct latent displacements, i.e., $B$ has full column rank.
The action that best explains a transition $(z, z')$ is then $G^{*}(z, z') = B^{+}(z' - Az)$, where $B^{+}$ is the pseudo-inverse of $B$; we refer to $G^{*}$ as the ideal IDM.

The realizability error splits into two parts, $e_t = e_t^{\sparallel} + e_t^{\perp}$.
The first part, which we call the \emph{action-explainable error}, is an error that some action could have produced. For example, if $a_t$ moves the agent to the right but the prediction shows it moving left, the predicted step looks exactly like the outcome of a different action.
The second part is an error that no action can produce, such as an object that suddenly appears in the scene.
Formally, $e_t^{\sparallel}$ is the orthogonal projection of $e_t$ onto $\mathrm{range}(B)$, the set of latent displacements that actions can cause, and $e_t^{\perp} = e_t - e_t^{\sparallel}$ (Appendix~\ref{sec:supp_theory_setting}).

\begin{proposition}[Residual as realizability error]
\label{prop:residual}
Let $f(z, a) = Az + Ba$, where $B$ has full column rank, and let $s_{\min}(B)$ and $s_{\max}(B)$ be the smallest and largest singular values of $B$.
Then, for every predicted transition $\hat{z}_{t+1} = F_\theta(\hat{z}_t, a_t)$:
\begin{enumerate}[label=(\roman*), leftmargin=*, itemsep=0pt, topsep=2pt]
    \item the residual of the ideal IDM is the realizability error mapped into the action space,
    \begin{equation}
    \label{eq:residual-identity}
    a_t - G^{*}\big(\hat{z}_t, \hat{z}_{t+1}\big) \;=\; -\,B^{+} e_t \;=\; -\,B^{+} e_t^{\sparallel} ;
    \end{equation}
    \item the residual bounds the action-explainable error from both sides,
    \begin{equation}
    \label{eq:two-sided}
    s_{\min}(B)\, \big\lVert B^{+} e_t \big\rVert \;\leq\; \big\lVert e_t^{\sparallel} \big\rVert \;\leq\; s_{\max}(B)\, \big\lVert B^{+} e_t \big\rVert .
    \end{equation}
\end{enumerate}
\end{proposition}

We prove Proposition~\ref{prop:residual} in Appendix~\ref{sec:supp_theory_identity} and~\ref{sec:supp_theory_proxy}.
By (i), the residual is the action-explainable error translated into the action space: it is the additional action that would have been needed to produce the prediction.
In the example of $e_t^{\sparallel}$, the IDM recovers a leftward action, and the residual is its gap from the commanded rightward action.
The action consistency cost is therefore not a heuristic that merely correlates with realizability; it measures the action-explainable part of the world model's own error.
By (ii), the residual bounds the action-explainable error from both sides: a large residual implies a large error, and a small residual certifies a small one.
The residual does not respond to $e_t^{\perp}$, so an object appearing from nowhere leaves it largely unchanged; for such errors, the augmented cost falls back to ranking candidates by $c_g$ alone.
Finally, the residual checks each step separately, whereas the goal cost observes only $\hat{z}_H$, into which every per-step error has already been propagated.

\textbf{Implementation of the inverse dynamics verifier.}
We instantiate the IDM, $G_\phi$, as a flow-matching~\citep{lipman2022flow, liu2022rectified} action decoder built on a compact prefix-suffix transformer following $\pi_0$~\citep{black2024pi_0}.
Given two consecutive latents $z_t$ and $z_{t+1}$, the model learns a velocity field $v_\phi(x_\tau, \tau, z_t, z_{t+1})$ that transports Gaussian noise $x_1 \sim \mathcal{N}(0, I)$ to the action $x_0 = a_t$ along the straight-line path $x_\tau = \tau\, \epsilon + (1 - \tau)\, a_t$.
Trained on real transitions, $G_\phi$ only approximates the ideal IDM $G^{*}$, and its approximation error enters the residual (Appendix~\ref{sec:supp_theory_learned}).
\Secref{sec:exp_idm_degrad} shows empirically that \modelname tolerates substantial IDM error.
Note that training $G_\phi$ requires no extra data, directly reusing training data for the world model. Also,
the verifier is lightweight at inference: a single Euler step suffices in our experiments without measurable degradation, adding negligible overhead to the planning loop.
We defer architectural and training details to Appendix~\ref{sec:supp_idm}.

\begin{algorithm}[t]
\caption{Decision-time Planning with Cycle Action Consistency}
\label{alg:cem-ours}
\begin{algorithmic}[1]
\small
\Require Encoder $E_\theta$, world model $F_\theta$, inverse dynamics $G_\phi$, current observation $o_0$, goal observation $o_g$, planning horizon $H$, \# of samples $N$, \# of elites $K$, \# of iterations $J$, consistency weight $\lambda$
\State Encode states: $z_0 \gets E_\theta(o_0)$,\ \ $z_g \gets E_\theta(o_g)$
\State Initialize sampling distribution $\mathcal{N}(\mu_0, \Sigma_0)$ with $\mu_0 = \mathbf{0}$, $\Sigma_0 = I$
\For{$j = 1$ to $J$}
    \State Sample $N$ action sequences $\{a^{(n)}_{0:H-1}\}_{n=1}^{N} \sim \mathcal{N}(\mu_{j-1}, \Sigma_{j-1})$
    \For{$n = 1$ to $N$}
        \State $\hat{z}_0^{(n)} \gets z_0$
        \For{$t = 0$ to $H-1$}
            \State $\hat{z}_{t+1}^{(n)} \gets F_\theta(\hat{z}_t^{(n)}, a^{(n)}_t)$ \Comment{World model predicted latent state}
            \State $\hat{a}_t^{(n)} \gets G_\phi(\hat{z}_t^{(n)}, \hat{z}_{t+1}^{(n)})$ \Comment{IDM inferred action}
        \EndFor
        \State Goal cost $c_g^{(n)} \gets \big\lVert \hat{z}_H^{(n)} - z_g \big\rVert_2^2$
        \State Action consistency cost $c_a^{(n)} \gets \frac{1}{H}\sum_{t=0}^{H-1}\big\lVert a^{(n)}_t - \hat{a}_t^{(n)}\big\rVert_2^2$
    \EndFor
    \State Adaptive weight $w_a \gets \lambda \cdot \sigma_g / \sigma_a$, where $\sigma_g \gets \mathrm{std}_n\!\big(c_g^{(n)}\big)$,\ \ $\sigma_a \gets \mathrm{std}_n\!\big(c_a^{(n)}\big)$
    \State Augmented planning cost $c^{(n)} \gets c_g^{(n)} + w_a\, c_a^{(n)}$ \ for $n = 1, \ldots, N$
    \State Select the $K$ sequences with lowest $c^{(n)}$ as elites $\mathcal{E}:$ $\mu_j \gets \frac{1}{K} \sum_{n \in \mathcal{E}} a^{(n)}_{0:H-1}$,  $\Sigma_j \gets \mathrm{Var}_{n \in \mathcal{E}}\big(a^{(n)}_{0:H-1}\big)$
\EndFor
\State \Return optimized action sequence $\mu_J$
\end{algorithmic}
\expandafter\gdef\csname @fs@post\endcsname{\kern2pt\hrule\kern-4mm\relax}%
\end{algorithm}
\expandafter\gdef\csname @fs@post\endcsname{\kern2pt\hrule\relax}%

\subsection{Planning with cycle action consistency}
\label{sec:planning}

\textbf{Augmented planning cost.}
Since the goal cost $c_g$ alone cannot penalize 
a candidate action sequence whose predicted trajectory is not realizable in the environment (\Secref{sec:act_con}), we augment the planning cost with the action consistency cost $c_a$ of Eq.~(\ref{eq:cyc-cost}).
The augmented
cost is
\begin{equation}
\label{eq:augmented-cost}
c(a_{0:H-1}) \;=\; c_g(a_{0:H-1}) \;+\; w_a \cdot c_a(a_{0:H-1}),
\end{equation}
and we use this cost in place of $c_g$ for CEM; the resulting procedure is summarized in Algorithm~\ref{alg:cem-ours},
with the consistency weight $w_a$ set adaptively as will be described next. 
This augmented cost encourages the planner to prioritize sequences whose predicted trajectory both \emph{reaches the goal} and \emph{remains realizable at every step}.
Under the linear model of \secref{sec:act_con} with the ideal IDM, the agent can end up far from the goal either because the predicted final state $\hat{z}_H$ is far from it, which $c_g$ measures, or because $\hat{z}_H$ is far from the state actually reached.
This latter gap accumulates the realizability errors of all steps, whose action-explainable parts $c_a$ bounds by Proposition~\ref{prop:residual}(ii).
Accordingly, for any $w_a > 0$ the true goal distance is at most a fixed multiple of $\sqrt{c_g + w_a c_a}$, plus a term from errors that no action can produce (Proposition~\ref{prop:goal-bound}, Appendix~\ref{sec:supp_theory_bound}); minimizing the augmented cost thus minimizes this bound, up to that term.
Because the augmentation changes only the per-candidate cost, it is by construction compatible with any search-based planning optimizer ({\it e.g.}, gradient- or sampling-based);
we instantiate it with CEM throughout.

\textbf{Scale-invariant adaptive weight.}
CEM ranks candidate action sequences according to their augmented costs, and selects those with the lowest costs, {\it i.e.}, elite candidates. 
Here, both the goal cost $c_g$ and the action consistency cost $c_a$ jointly determine the ranking of candidates,
so the relative influence of the two costs decides which candidates become elites.
This 
ranking
is governed not by the absolute magnitudes of the two costs, but by how spread out their relative values across the candidates. 
In other words, if a cost barely varies across candidates, it will contribute little to their ranks regardless of its scale. 
We measured each cost's spread by its standard deviation over the $N$ candidates, $\sigma_g$ for $c_g$ and $\sigma_a$ for $c_a$, and empirically observed that it varies substantially across (1) world models, since different latent spaces produce inherently different cost distributions, (2) tasks, since the magnitude of visual change between frames differs by task and shifts the scale of the cost standard deviations, and even (3) CEM iterations, since the two spreads compress at different rates when the sampling distribution concentrates around the elites; we provide a detailed empirical analysis of these variations in Appendix~\ref{sec:supp_weight}.
A fixed value for $w_a$ thus generalizes poorly as it lets the impact of the action consistency cost drift in the ranking process.
To ensure the two costs demonstrate comparable influence on elite selection regardless of these variations, we set the weight adaptively at every CEM iteration:
\begin{equation}
\label{eq:idm-weight}
w_a \;=\; \lambda \cdot \frac{\sigma_g}{\sigma_a},
\end{equation}
which equalizes the spread of the consistency cost and the goal cost up to the factor $\lambda$, with $\lambda$ controlling which term to prioritize during planning.
In practice, $\lambda$ needs no per-task retuning: for each world model, a single value is shared across all tasks of a benchmark suite.

\section{Experiment}
\label{sec:experiment}

Our experiments ask whether augmenting decision-time planning with cycle action consistency improves control, how robust the gain is to the planner and to the verifier, and at what cost. We organize the study around four questions.
\textbf{Q1) Generality:} Does cycle action consistency improve planning across world models with fundamentally different latent spaces, across simulated tasks, and on a real robot (\secref{sec:exp_main} and \secref{sec:real_robot})?
\textbf{Q2) Hyperparameter robustness:} Does the gain hold across CEM planning budgets and consistency weights $\lambda$ without per-task tuning (\secref{subsec:lambda_exp})?
\textbf{Q3) Net efficiency:} Despite the per-iteration overhead of the inverse dynamics verifier, does \modelname{} reach a target quality with less total planning compute (\secref{sec:exp_efficiency})?
\textbf{Q4) Robustness to verifier error:} How much error in the inverse dynamics verifier can the gain tolerate (\secref{sec:exp_idm_degrad})?

\subsection{Experimental settings}
\begin{figure}[t!]
    \centering
    \includegraphics[width=1.00\columnwidth]{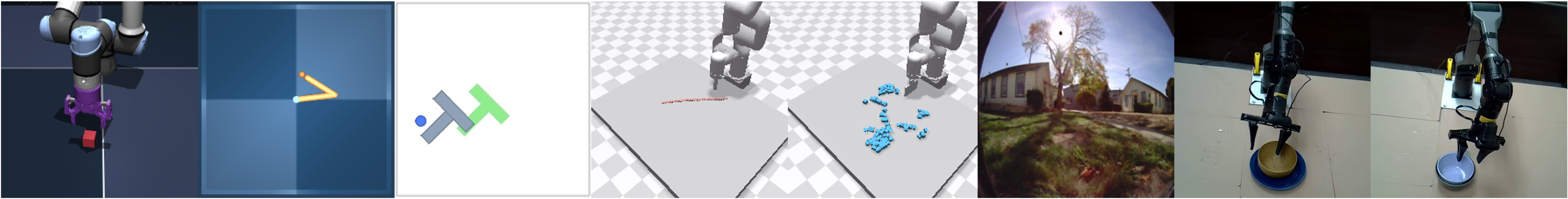}
    \vspace{-6mm}
    \caption{\textbf{Environment suites used in our experiments.} From left to right: Cube, Reacher, PushT, Rope Manipulation, Granular Manipulation, goal-conditioned visual navigation on RECON~\citep{shah2021rapid}, real robot tasks (PickPlace \& BowlStack). More details on the environments and datasets are available in Appendix~\ref{sec:supp_exp_settings}. Results on visual navigation are presented in Appendix~\ref{sec:supp_nwm}.}
    \label{fig:env}
\vspace{-5mm}
\end{figure}

\noindent \textbf{Environments and tasks.}
We evaluate on six tasks spanning a broad range of dynamics (\Figref{fig:env}): robotic arm manipulation (Cube), articulated control (Reacher), fine-grained tabletop pushing (PushT), deformable object manipulation (Rope and Granular), and goal-conditioned visual navigation (Appendix~\ref{sec:supp_nwm}). For the real robot, we evaluate on two tabletop tasks (PickPlace and BowlStack).

\noindent \textbf{World model baselines.}
We build planning on top of three JEPA-style latent predictors that differ in how their latent spaces are acquired.
(1) DINO-WM~\citep{zhou2024dino} learns the predictor over frozen features from a pretrained DINOv2 encoder, while (2) PLDM~\citep{sobal2026learning} and (3) Le-WM~\citep{maes2026leworldmodel} instead learn the encoder and predictor jointly from pixels, preventing collapse with a VICReg-family objective and a single normality regularizer, respectively.
We evaluate all three world models on all five simulated tasks: Cube, Reacher, PushT, Rope, and Granular. We additionally apply \modelname{} to a video generation-based world model, NWM trained with CompACT~\citep{bar2025navigation,kim2026planning}, on goal-conditioned visual navigation (Appendix~\ref{sec:supp_nwm}).
For the real robot experiment, we use a separate JEPA-style latent predictor.

\noindent \textbf{Evaluation.}
For Cube, Reacher, and PushT, we report success rate (\%), the fraction of episodes whose rollout reaches the goal pose within a task-specific tolerance. 
For Rope and Granular, where success is not naturally binary, we instead report Chamfer distance between the achieved and target configurations.
For the real robot tasks, we report success rate as the number of successful episodes over the total number of evaluation episodes.

\subsection{Planning across diverse world models and environments}
\label{sec:exp_main}
\begin{table}[t]
\centering
\caption{\textbf{Planning performance across world models and tasks.} The metric is success rate (\%, higher is better) on Cube, Reacher, and PushT, and Chamfer distance (lower is better) on Rope and Granular. \emph{Original} plans with the goal cost only, while \emph{Ours} adds the action consistency cost. Values are mean $\pm$ std over three seeds.}
\vspace{-3mm}
\label{tab:main}
\scalebox{0.85}{
\begin{tabular}{llccccc}
\toprule
\multirow{2}{*}{Model} & \multirow{2}{*}{Planning cost} & Cube & Reacher & PushT & Rope & Granular \\
\cmidrule(lr){3-5} \cmidrule(lr){6-7}
 & & \multicolumn{3}{c}{Success rate (\%) $\uparrow$} & \multicolumn{2}{c}{Chamfer distance $\downarrow$} \\
\midrule
\multirow{2}{*}{Le-WM}   & Original & 68.0\,{\scriptsize$\pm$\,5.3} & 75.3\,{\scriptsize$\pm$\,2.3} & 92.7\,{\scriptsize$\pm$\,3.1} & 1.79\,{\scriptsize$\pm$\,0.07} & 0.59\,{\scriptsize$\pm$\,0.02} \\
                         & Ours     & \textbf{79.3}\,{\scriptsize$\pm$\,3.1} & \textbf{82.0}\,{\scriptsize$\pm$\,2.0} & \textbf{94.0}\,{\scriptsize$\pm$\,2.0} & \textbf{1.26}\,{\scriptsize$\pm$\,0.21} & \textbf{0.43}\,{\scriptsize$\pm$\,0.04} \\
\midrule
\multirow{2}{*}{PLDM}    & Original & 56.0\,{\scriptsize$\pm$\,2.0} & 75.3\,{\scriptsize$\pm$\,3.1} & 74.7\,{\scriptsize$\pm$\,3.1} & 1.95\,{\scriptsize$\pm$\,0.10} & 0.60\,{\scriptsize$\pm$\,0.09} \\
                         & Ours     & \textbf{68.0}\,{\scriptsize$\pm$\,2.0} & \textbf{82.7}\,{\scriptsize$\pm$\,3.1} & \textbf{78.7}\,{\scriptsize$\pm$\,1.2} & \textbf{1.38}\,{\scriptsize$\pm$\,0.03} & \textbf{0.49}\,{\scriptsize$\pm$\,0.09} \\
\midrule
\multirow{2}{*}{DINO-WM} & Original & 76.7\,{\scriptsize$\pm$\,2.3} & 67.3\,{\scriptsize$\pm$\,3.1} & 75.3\,{\scriptsize$\pm$\,6.1} & 1.32\,{\scriptsize$\pm$\,0.08} & 0.46\,{\scriptsize$\pm$\,0.02} \\
                         & Ours     & \textbf{81.3}\,{\scriptsize$\pm$\,6.1} & \textbf{68.0}\,{\scriptsize$\pm$\,5.3} & \textbf{76.7}\,{\scriptsize$\pm$\,6.1} & \textbf{0.65}\,{\scriptsize$\pm$\,0.07} & \textbf{0.34}\,{\scriptsize$\pm$\,0.04} \\
\bottomrule

\end{tabular}
}
\vspace{-2mm}
\end{table}

\begin{figure}[t!]
    \centering
    \includegraphics[width=\linewidth]{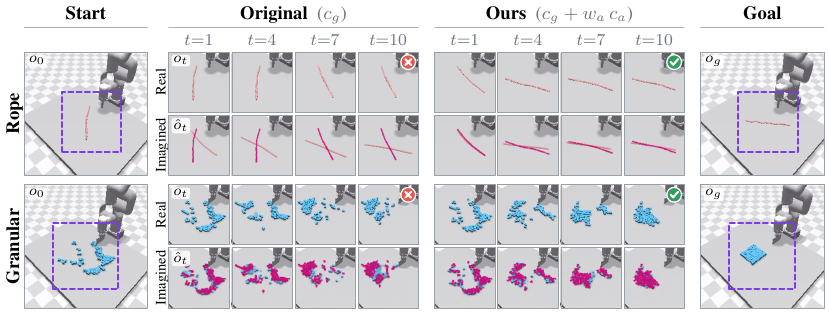}
    \vspace{-6mm}
    \caption{\textbf{Qualitative comparison on Rope and Granular with DINO-WM.} From the start observation $o_0$ (dashed box: zoomed region), we show the environment rollout $o_t$ (top row) and the world model's imagined rollout $\hat o_t$ (bottom row) of the planned action sequence, with the real object overlaid on $\hat o_t$ in magenta. Under \emph{Original}, the imagined rollout reaches the goal $o_g$ while the real one does not; with the action consistency cost (\emph{Ours}), the two coincide and the goal is reached.
}
    \label{fig:qual_rope_granular}
\vspace{-5mm}
\end{figure}

Each of the three world models we consider plans with its own goal cost $c_g$ of Eq.~(\ref{eq:latent-cost}), without any modification to the planning cost.
We thus adopt them as baselines and compare planning with and without the proposed action consistency cost. 
For every model we keep its training and the underlying CEM planner fixed, and vary only the planning cost.
Throughout, \emph{Original} plans with the goal cost only, and \emph{Ours} adds the action consistency cost.

\noindent \textbf{Planning performance.}
Table~\ref{tab:main} demonstrates that incorporating the proposed action consistency cost improves planning across every world model and task.
For the JEPA-style latent predictors, Le-WM and PLDM, success rate improves on all of Cube, Reacher, and PushT.
On Rope and Granular manipulation tasks, Chamfer distance decreases substantially for all three world models, in particular DINO-WM on Rope.
We attribute these consistent gains to a mechanism that does not depend on the task or the world model: the model can produce predicted trajectories that no action sequence could actually realize, and the action consistency cost suppresses these non-realizable candidates by checking each against the actions an inverse dynamics model infers from it.
The action consistency cost modifies only the planning cost and leaves the world model untouched, so it is orthogonal to world model improvements and can be composed with any action-conditioned world model.
This extends to video generation-based world models: on goal-conditioned visual navigation with NWM trained with CompACT, the action consistency cost 
yields notable performance improvement (Appendix~\ref{sec:supp_nwm}).

\noindent \textbf{Qualitative results.}
As shown in \Figref{fig:qual_rope_granular},
the complex dynamics of deformable and granular media make many CEM candidate trajectories non-realizable in the actual environment, precisely the failure mode the action consistency cost is designed to cope with.
For the same episode, we contrast the world model's imagined rollout with the real  rollout obtained by executing the planned actions.
Under the \emph{Original} planning cost, the planned action sequence reaches the goal in the imagined rollout but drifts away from it in the real rollout, leaving the goal cost satisfied by an unreachable future.
Adding the action consistency cost (\emph{Ours}) removes this discrepancy: the real rollout closely tracks the imagined rollout and reaches the goal, confirming that the consistency term filters out the unrealizable candidates the goal cost alone cannot detect.

\subsection{Real robot experiment}
\label{sec:real_robot}
On a real robot, sampling joint-space action sequences from scratch is inefficient, so we adopt the policy-guided planning protocol of \citet{chun2026sparse}, in which the world model selects among action chunks proposed by a learned policy.
Since \modelname{} changes only the per-candidate cost, it applies to this setting as is.

\noindent \textbf{Setup.}
At each decision step, we sample $N=20$ action chunks from the policy, predict one world-model step ahead for each, and execute the candidate with the lowest planning cost.
The policy is a Diffusion Policy~\citep{chi2023diffusionpolicy}, the world model is a JEPA-style latent predictor trained with the temporal straightening regularizer of \citet{wang2026temporal}, and the IDM is trained on its frozen latents, all from scratch on teleoperation demonstrations.
We evaluate on two tabletop tasks using a PiPER arm with a single RGB camera: PickPlace, which places a bowl on a plate, and BowlStack, which stacks one bowl on top of another. 
We run one episode from each of 15 initial bowl locations unseen during training; more details are available in Appendix~\ref{sec:supp_real_robot}.

\begin{wraptable}{r}{0.45\textwidth}
\centering
\vspace{-5mm}
\caption{\textbf{Policy-guided planning performance on the real robot.} Success rate over 15 initial bowl positions per task.}
\vspace{-2.5mm}
\label{tab:real_robot}
\scalebox{0.7}{
\begin{tabular}{llcc}
\toprule
Model & Planning cost & PickPlace & BowlStack \\
\midrule
Diffusion policy & -- & 7/15 & 8/15 \\
\midrule
$+$World model & Original & 11/15 & 11/15 \\
$+$IDM         & Ours     & \textbf{14/15} & \textbf{14/15} \\
\bottomrule
\end{tabular}
}
\vspace{-3mm}
\end{wraptable}
\noindent \textbf{Planning performance.}
As shown in Table~\ref{tab:real_robot}, reranking the policy's samples by the goal cost alone (\emph{Original}) raises success over a single policy sample from 7/15 to 11/15 on PickPlace and from 8/15 to 11/15 on BowlStack, and adding the action consistency cost (\emph{Ours}) raises it further to 14/15 on both tasks.
Since the policy, the world model, and the candidates are identical between the two, this gain comes from the planning cost alone, and \modelname{} improves policy-guided planning without retraining either model.

\begin{table}[h!]
\centering
\begin{minipage}[t]{0.45\textwidth}
\centering
\caption{\textbf{CEM budget and $\lambda$ sweeps on Reacher.}
(\textit{Left}): CEM budget sweep with $\lambda = 0.07$.
(\textit{Right}): $\lambda$ sweep with $N = 300$.}
\vspace{-3mm}
\label{tab:robustness}
\setlength{\tabcolsep}{3pt}
\scalebox{0.7}{
\begin{tabular}{ll|cccc|cccc}
\toprule
\multirow{2}{*}{Model} & \multirow{2}{*}{\makecell[l]{Planning \\ cost}} & \multicolumn{4}{c|}{\# CEM samples ($N$)} & \multicolumn{4}{c}{$\lambda$} \\
            &               & 30 & 50 & 150 & 300 & 0.005 & 0.04 & 0.07 & 0.1 \\
\midrule
\multirow{2}{*}{Le-WM} & Original & 68 & 62 & 70 & 76 & \multicolumn{4}{c}{76} \\
                       & Ours     & \textbf{82} & \textbf{76} & \textbf{78} & \textbf{88} & \textbf{88} & \textbf{80} & \textbf{88} & \textbf{90} \\
\midrule
\multirow{2}{*}{PLDM}  & Original & 58 & 66 & 70 & 76 & \multicolumn{4}{c}{76} \\
                       & Ours     & \textbf{78} & \textbf{84} & \textbf{82} & \textbf{84} & \textbf{90} & \textbf{92} & \textbf{84} & \textbf{88} \\
\bottomrule
\end{tabular}
}
\end{minipage}
\hfill
\begin{minipage}[t]{0.5\textwidth}
\centering
\caption{\textbf{Constant vs.\ scale-invariant adaptive weight $w_a$ on Le-WM.} Cells show the change in success rate ($\Delta$, \%) over \emph{Original}.
}
\vspace{-3mm}
\label{tab:constant_idm_weight}
\setlength{\tabcolsep}{4pt}
\scalebox{0.7}{
\begin{tabular}{lc|ccccc|c}
\toprule
\multirow{2}{*}{Task} & \multirow{2}{*}{\makecell{Original}} & \multicolumn{5}{c|}{Constant $w_a$} & \multirow{2}{*}{Ours} \\
 & & 1.0 & 3.0 & 5.0 & 7.0 & 10.0 & \\
\midrule
Cube    & 70.0 & $+0.0$ & $-2.0$ & $+0.0$ & $+0.0$ & $+2.0$ & $+4.0$ \\
PushT   & 96.0 & $+0.0$ & $+2.0$ & $-2.0$ & $+0.0$ & $+2.0$ & $+4.0$ \\
Reacher & 76.0 & $+14.0$ & $+6.0$ & $-2.0$ & $+4.0$ & $+6.0$ & $+12.0$ \\
\midrule
Total $\Delta$ & -- & $+14.0$ & $+6.0$ & $-4.0$ & $+4.0$ & $+10.0$ & $\mathbf{+20.0}$ \\
\bottomrule
\end{tabular}
}
\end{minipage}
\vspace{-3mm}
\end{table}

\subsection{Robustness to CEM planning budget and $\lambda$}
\label{subsec:lambda_exp}

In Table~\ref{tab:robustness}(left), we vary the number of CEM samples from $30$ to $300$ while keeping top-$K$ at $10\%$.
\emph{Ours} consistently outperforms \emph{Original} at every budget for both world models, never falling below the baseline.
Table~\ref{tab:robustness}(right) sweeps the consistency weight $\lambda$ over $4$ values spanning a $20\times$ range ($0.005$ to $0.1$).
The results indicate that a broad band of $\lambda$ yields consistent gains: within each model the method does not require careful tuning of $\lambda$ per task to be effective.
\emph{Ours} exhibits this level of robustness thanks to the scale-invariant adaptive weight proposed in Sec.~\ref{sec:planning}:
Table~\ref{tab:constant_idm_weight} isolates why the weight should be adaptive rather than constant.
The best constant $w_a$ differs across tasks, which a fixed weight cannot accommodate.
Our scale-invariant adaptive weight $w_a = \lambda \cdot \sigma_g / \sigma_a$, set per CEM iteration, removes this dependence and yields the largest total improvement, the robustness the sweeps above reflect.
\subsection{Planning compute analysis}
\label{sec:exp_efficiency}

\begin{figure}[t]
    \centering
    \begin{minipage}[t]{0.565\textwidth}
        \centering
        \includegraphics[width=\linewidth]{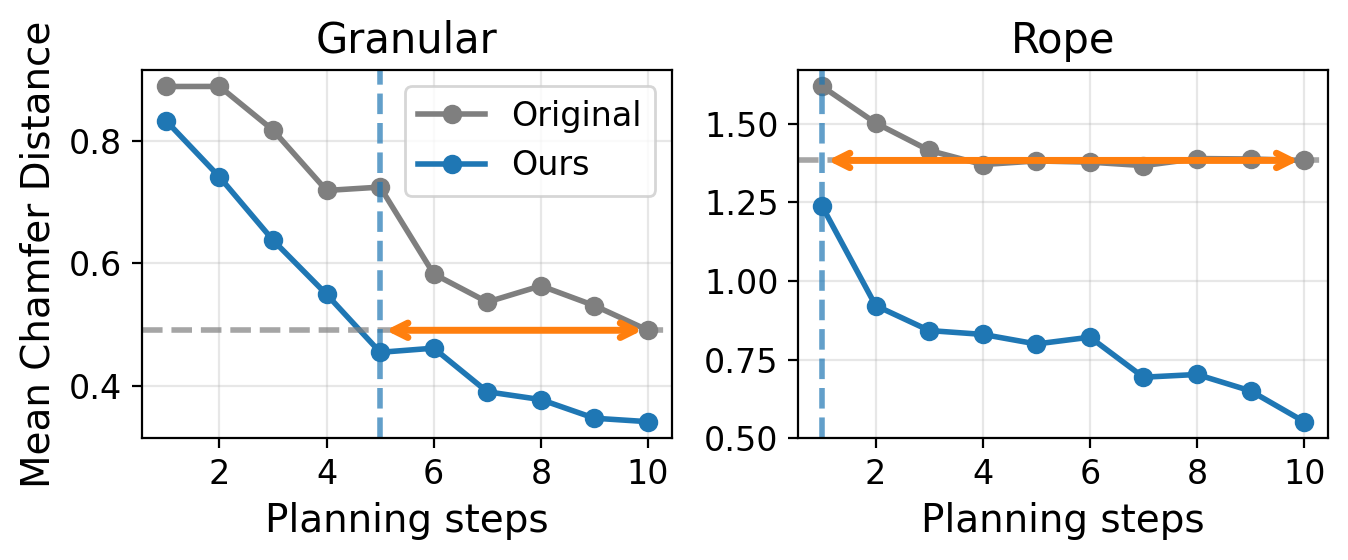}
        \vspace{-7mm}
        \caption{\textbf{Mean Chamfer distance over planning steps on Granular and Rope with DINO-WM} (lower is better). The horizontal dashed line marks the Chamfer distance that \emph{Original} reaches at its final step, and the vertical dashed line marks the step where \emph{Ours} first reaches it.}
        \label{fig:qual_chamfer}
    \end{minipage}\hfill
    \begin{minipage}[t]{0.41\textwidth}
        \centering
        \includegraphics[width=\linewidth]{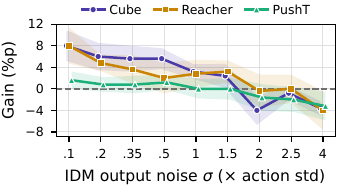}
        \vspace{-7mm}
        \caption{\textbf{Gain in success rate over \emph{Original} under increasing IDM error, with PLDM.} $\sigma$ is the injected error in units of the dataset's action standard deviation; drawn over $5$ seeds per task.}
        \label{fig:idm_degrad}
    \end{minipage}
\vspace{-5mm}
\end{figure}

\noindent \textbf{Per-iteration overhead of the inverse dynamics verifier.}
We measure the verifier's per-iteration overhead, defined as the ratio of the IDM's forward time to the world model's forward time within one CEM iteration. The IDM is run with a single integration step on every predicted transition of every candidate, while the world model rolls out the same candidates over the full horizon.
The relative overhead is 14.5\%, 14.5\%, and 11.8\% for Le-WM, PLDM, and DINO-WM, respectively; exact latencies are reported in Appendix~\ref{sec:supp_latency}.

\noindent \textbf{Lower total compute to a target quality.}
Since total planning compute is proportional to both the number of CEM samples and the number of planning steps, reducing either factor directly reduces the overall cost.
We set the target to the quality that \emph{Original} attains with its full budget and measure how much of that budget \emph{Ours} requires.
For Le-WM and PLDM on Reacher, the saving appears along the sample axis: in the CEM-budget sweep of Table~\ref{tab:robustness}, \emph{Ours} with the smallest budget ($N{=}30$) already exceeds \emph{Original} with the full budget ($N{=}300$), using one tenth of the samples.
For DINO-WM on Granular and Rope, it appears along the planning-step axis (\Figref{fig:qual_chamfer}): \emph{Ours} reaches the Chamfer distance that \emph{Original} attains only at its final planning step in less than half the planning steps on Granular, and already after the first planning step on Rope.
Thus, the samples and planning steps \emph{Ours} saves in reaching the target more than compensate for the per-iteration overhead the verifier imposes.

\subsection{Robustness to inverse dynamics verifier error}
\label{sec:exp_idm_degrad}

\modelname{} uses a learned IDM rather than the oracle of our analysis, so the actions it infers carry an error; Appendix~\ref{sec:supp_theory_learned} argues that this error does not systematically mask world-model errors.
We test how much IDM error \modelname{} tolerates by adding, at each replanning step, a random Gaussian offset of scale $\sigma$ to the IDM's predictions, where $\sigma$ is in units of the dataset's action standard deviation.
We sweep $\sigma$ from $0.1$ to $4$ with PLDM on Cube, Reacher, and PushT.
As \Figref{fig:idm_degrad} shows, the gain over \emph{Original} shrinks as $\sigma$ grows, but it remains non-negative on all three tasks up to $\sigma = 1.5$, an error larger than the spread of the actions themselves, and turns negative only from $\sigma = 2$ onward.

\section{Conclusion}
\label{sec:conclusion}

We introduced \modelname{}, a decision-time planning framework that augments the planning cost of action-conditioned world models with \emph{cycle action consistency}.
A separately trained inverse dynamics model maps each predicted transition back to the action space, and its residual against the conditioning action, an estimate of the world model's rollout error, is added to the standard goal cost.
This consistently improves planning across diverse world models and tasks, spanning rigid and deformable object manipulation, articulated control, and visual navigation, and carries over to a real robot arm.
Since the new cost changes only the per-candidate score, it is agnostic to the optimizer; while we evaluate it with sampling-based planners, extending it to gradient-based planning and composing it with stronger world-model backbones are natural next steps.

\clearpage

\bibliography{main}  %
\bibliographystyle{iclr2027_conference}
\clearpage

\setcounter{section}{0}
\renewcommand{\thesection}{\Alph{section}}
\section*{Appendix}
\renewcommand{\thesection}{\Alph{section}}
\renewcommand{\theHsection}{appendix.\Alph{section}}
\section{Theoretical analysis of cycle action consistency}
\label{sec:supp_theory}

In Sec.~\ref{sec:act_con} we introduced the action consistency cost $c_a$ as a signal that detects predicted transitions not realizable by their conditioning actions. Here we make this argument precise under locally linear latent dynamics. We show that (i)~the residual of an ideal inverse dynamics model is exactly the world model's one-step error mapped into the action space (Sec.~\ref{sec:supp_theory_identity}), (ii)~this residual bounds the action-explainable error from both sides (Sec.~\ref{sec:supp_theory_proxy}), and (iii)~the augmented cost of Eq.~(\ref{eq:augmented-cost}) upper-bounds the true distance to the goal, up to the part of the error that no action can produce (Sec.~\ref{sec:supp_theory_bound}). Throughout, the IDM enters only through a fixed map determined by the true dynamics and acts as an estimator of the world model's per-step rollout error.

\subsection{Setting and notation}
\label{sec:supp_theory_setting}

\noindent\textbf{Locally linear latent dynamics.}
We model the true latent dynamics $f$ as
\begin{equation}
\label{eq:lin-dyn}
z_{t+1} \;=\; f(z_t, a_t) \;=\; A z_t + B a_t,
\end{equation}
where $A \in \mathbb{R}^{d \times d}$ describes the passive drift of the latent state and $B \in \mathbb{R}^{d \times d_a}$, with $d_a < d$, maps an action to the latent displacement it causes. For the nonlinear world models in our experiments, Eq.~(\ref{eq:lin-dyn}) should be read as the linearization of the dynamics around the predicted state, so every constant below is local. A linear latent model of this form is also adopted by \citet{wang2026temporal}; we relax their invertibility assumption on $B$ to the following.

\begin{assumption}[Identifiability]
\label{assump:identifiability}
$B$ has full column rank.
\end{assumption}

Assumption~\ref{assump:identifiability} states that distinct actions always produce distinct latent displacements, so the action between two consecutive latent states is uniquely determined. This is the property \modelname exploits throughout (Sec.~\ref{sec:intro}). It concerns how actions register in the latent space rather than whether the dynamics are deterministic (Sec.~\ref{sec:supp_theory_learned}).

\noindent\textbf{Pseudo-inverse and singular values.}
We denote by $s_{\max}(B) \geq s_{\min}(B)$ the largest and smallest singular values of $B$; we write $s$ rather than $\sigma$ to avoid confusion with the cost standard deviations $\sigma_g$ and $\sigma_a$ of Eq.~(\ref{eq:idm-weight}). Assumption~\ref{assump:identifiability} is equivalent to $s_{\min}(B) > 0$. Under this assumption, the Moore--Penrose pseudo-inverse of $B$ is $B^{+} = (B^\top B)^{-1} B^\top \in \mathbb{R}^{d_a \times d}$, and $P = B B^{+}$ is the orthogonal projection onto $\mathrm{range}(B)$, the subspace of latent displacements that some action can produce. We use the following standard facts.

\begin{lemma}
\label{lem:pinv}
Under Assumption~\ref{assump:identifiability}, the following hold.
\begin{enumerate}[label=(\alph*), leftmargin=*, itemsep=0pt, topsep=2pt]
\item $B^{+} B = I_{d_a}$.
\item $B B^{+} = P$.
\item $B^{+} = B^{+} P$, so that $B^{+} v = 0$ for every $v \perp \mathrm{range}(B)$.
\item $s_{\min}(B)\,\lVert x \rVert \leq \lVert B x \rVert \leq s_{\max}(B)\,\lVert x \rVert$ for every $x \in \mathbb{R}^{d_a}$.
\end{enumerate}
\end{lemma}

\noindent\textbf{Ideal inverse dynamics model.}
Under Eq.~(\ref{eq:lin-dyn}), the action that explains a transition $(z, z')$ in the least-squares sense is
\begin{equation}
\label{eq:oracle-idm}
G^{*}(z, z') \;=\; \operatorname*{arg\,min}_{a}\, \big\lVert z' - A z - B a \big\rVert_2^2 \;=\; B^{+}\big(z' - A z\big).
\end{equation}
On every real transition, $z_{t+1} - A z_t = B a_t$, so Lemma~\ref{lem:pinv}(a) gives $G^{*}(z_t, z_{t+1}) = B^{+} B a_t = a_t$: the ideal IDM recovers the executed action exactly. We refer to $G^{*}$ as the oracle IDM. The learned verifier $G_\phi$ approximates it on the transitions of the offline dataset, and we account for its approximation error in Sec.~\ref{sec:supp_theory_learned}.

\noindent\textbf{Realizability error.}
We measure how far a predicted transition is from being realizable by comparing it with the transition that its conditioning action actually produces from the same state.

\begin{definition}[Realizability error]
\label{def:realizability-error}
For a predicted transition $\hat{z}_{t+1} = F_\theta(\hat{z}_t, a_t)$, the realizability error is
\begin{equation}
\label{eq:realizability-error}
e_t \;=\; \hat{z}_{t+1} - f(\hat{z}_t, a_t) \;=\; \hat{z}_{t+1} - A \hat{z}_t - B a_t.
\end{equation}
\end{definition}

The realizability error $e_t$ is the world model's one-step error at its own input: the predicted transition is realizable by its conditioning action if and only if $e_t = 0$. Because $e_t$ is measured from the predicted state $\hat{z}_t$, it excludes the drift accumulated at earlier steps, which is already contained in $\hat{z}_t$. We decompose it as $e_t = e_t^{\sparallel} + e_t^{\perp}$, where $e_t^{\sparallel} = P e_t \in \mathrm{range}(B)$ and $e_t^{\perp} = (I - P)\, e_t \perp \mathrm{range}(B)$. The component $e_t^{\sparallel}$ is a deviation that some other action could have produced, whereas $e_t^{\perp}$ is a deviation that no action can produce. \Figref{fig:supp_theory_geometry} illustrates this decomposition.

\begin{figure}[t]
    \centering
    \begin{tikzpicture}[scale=1.25, >=stealth]
        \draw[gray!70, line width=0.8pt] (-0.8,-0.4) -- (7.0,3.5) node[right, black] {$A\hat{z}_t + \mathrm{range}(B)$};
        \draw[->, line width=1.1pt, black, shorten >=1.6pt] (0,0) -- (2.4,1.2);
        \node[above left] at (1.2,0.6) {$B a_t$};
        \draw[->, dotted, line width=0.9pt, shorten >=2.4pt] (2.4,1.2) -- (4.2,3.4);
        \node[left] at (3.3,2.3) {$e_t$};
        \draw[->, line width=1.6pt, green!55!black, shorten >=2.2pt] (2.4,1.2) -- (4.72,2.36);
        \node[below right, green!45!black] at (3.56,1.78) {$e_t^{\sparallel} = B B^{+} e_t$};
        \draw[->, dashed, line width=1.1pt, magenta!80!black, shorten >=2.4pt] (4.72,2.36) -- (4.2,3.4);
        \node[right, magenta!80!black] at (4.46,2.88) {$e_t^{\perp}$};
        \draw[line width=0.5pt] (4.559,2.280) -- (4.479,2.441) -- (4.640,2.521);
        \fill (0,0) circle (1.6pt) node[below right] {$A\hat{z}_t$};
        \fill (2.4,1.2) circle (1.6pt) node[above left] {$f(\hat{z}_t, a_t)$};
        \draw[green!45!black, fill=white, line width=1pt] (4.72,2.36) circle (2.2pt);
        \node[below right, green!45!black] at (4.72,2.36) {$f(\hat{z}_t, a_t + B^{+}e_t)$};
        \fill[orange!85!black] (4.2,3.4) circle (2.4pt);
        \node[above, orange!85!black] at (4.2,3.45) {$\hat{z}_{t+1}$};
    \end{tikzpicture}
    \caption{\textbf{Geometry of the realizability error.} Dots are latent states and arrows are displacements. From $\hat{z}_t$, the passive drift reaches $A\hat{z}_t$, and the states that some action can reach form the line $A\hat{z}_t + \mathrm{range}(B)$. The conditioning action $a_t$ reaches $f(\hat{z}_t, a_t) = A\hat{z}_t + B a_t$, whereas the world model predicts $\hat{z}_{t+1}$, which is off by $e_t$. The oracle IDM projects $\hat{z}_{t+1}$ onto the line and reads off the action $a_t + B^{+} e_t$, which reaches the foot of the projection (Proposition~\ref{prop:residual}): the component $e_t^{\sparallel}$ shifts the recovered action, whereas $e_t^{\perp}$ is discarded.}
    \label{fig:supp_theory_geometry}
\end{figure}

\subsection{The residual is the world model's error in the action space}
\label{sec:supp_theory_identity}

Proposition~\ref{prop:residual} in Sec.~\ref{sec:act_con} states two properties of the residual, which we prove in this and the next subsection. Its full-column-rank condition is Assumption~\ref{assump:identifiability}, and $G^{*}$ is the ideal IDM of Eq.~(\ref{eq:oracle-idm}).

\begin{proof}[Proof of Proposition~\ref{prop:residual}(i)]
By Eq.~(\ref{eq:oracle-idm}), Eq.~(\ref{eq:realizability-error}), and Lemma~\ref{lem:pinv}(a),
$G^{*}(\hat{z}_t, \hat{z}_{t+1}) = B^{+}(\hat{z}_{t+1} - A \hat{z}_t) = B^{+}(B a_t + e_t) = a_t + B^{+} e_t$,
which rearranges to the first equality of Eq.~(\ref{eq:residual-identity}). The second follows from Lemma~\ref{lem:pinv}(c).
\end{proof}

Substituting Eq.~(\ref{eq:residual-identity}) into Eq.~(\ref{eq:cyc-cost}) shows that, with the ideal IDM, the action consistency cost is
\begin{equation}
\label{eq:ca-oracle}
c_a(a_{0:H-1}) \;=\; \frac{1}{H} \sum_{t=0}^{H-1} \big\lVert B^{+} e_t \big\rVert_2^2 .
\end{equation}

Proposition~\ref{prop:residual}(i) states that the IDM reports the conditioning action shifted by $\Delta a_t = B^{+} e_t$, the action-space image of the world model's error; $\Delta a_t$ is the additional action that would explain the predicted displacement. The residual therefore measures the world model's own one-step error, and the IDM acts as an estimator of this error through the fixed map $B^{+}$, a function of the true dynamics. Moreover, $B^{+} e_t$ depends only on the error made at step $t$ and does not compound along the horizon, whereas the goal cost $c_g$ observes only $\hat{z}_H$, into which the world model has already propagated every $e_0, \ldots, e_{H-1}$ (Sec.~\ref{sec:supp_theory_bound}).

By Lemma~\ref{lem:pinv}(c), $B^{+} e_t = B^{+} e_t^{\sparallel}$. The residual thus vanishes if and only if $e_t \perp \mathrm{range}(B)$: it vanishes whenever the prediction is realizable ($e_t = 0$), but it does not detect a deviation that lies entirely outside $\mathrm{range}(B)$. We return to this blind spot in Sec.~\ref{sec:supp_theory_bound}.

\subsection{The residual is a two-sided proxy of the actionable error}
\label{sec:supp_theory_proxy}

The residual is measured in the action space, while the error it reflects lives in the latent space. Proposition~\ref{prop:residual}(ii) relates their magnitudes.

\begin{proof}[Proof of Proposition~\ref{prop:residual}(ii)]
By Lemma~\ref{lem:pinv}(b), $e_t^{\sparallel} = P e_t = B B^{+} e_t = B\,\Delta a_t$. Applying Lemma~\ref{lem:pinv}(d) with $x = \Delta a_t$ gives $s_{\min}(B)\,\lVert \Delta a_t \rVert \leq \lVert e_t^{\sparallel} \rVert \leq s_{\max}(B)\,\lVert \Delta a_t \rVert$, which is Eq.~(\ref{eq:two-sided}).
\end{proof}

The two inequalities of Eq.~(\ref{eq:two-sided}) play different roles. The right inequality, $\lVert e_t^{\sparallel} \rVert \leq s_{\max}(B)\,\lVert B^{+} e_t \rVert$, states that the scaled residual never underestimates the action-explainable error: a small residual certifies a small $e_t^{\sparallel}$, so such an error cannot escape the cost. Pessimistic model-based reinforcement learning requires exactly such an upper bound on the model's error~\citep{yu2020mopo}, and here it holds by construction. The left inequality, $s_{\min}(B)\,\lVert B^{+} e_t \rVert \leq \lVert e_t^{\sparallel} \rVert$, states that the residual cannot be large unless the error is, so an accurate prediction is never penalized heavily. It relies on Assumption~\ref{assump:identifiability} and weakens as $s_{\min}(B) \to 0$. Both bounds are attained, for errors aligned with the left singular vectors of $B$ associated with $s_{\max}(B)$ and $s_{\min}(B)$, respectively.

\subsection{The augmented cost bounds the true goal distance}
\label{sec:supp_theory_bound}

The goal cost $c_g$ of Eq.~(\ref{eq:latent-cost}) scores a candidate by its predicted terminal state alone. We now relate the augmented cost of Eq.~(\ref{eq:augmented-cost}) to the terminal state $z_H$ reached when the same action sequence is executed from $z_0$ under the true dynamics. Let $\delta_t = \hat{z}_t - z_t$ denote the drift of the predicted trajectory from the true one, with $\delta_0 = 0$. Subtracting $z_{t+1} = A z_t + B a_t$ from $\hat{z}_{t+1} = A \hat{z}_t + B a_t + e_t$ gives
\begin{equation}
\label{eq:drift}
\delta_{t+1} \;=\; A\,\delta_t + e_t,
\qquad\textup{hence}\qquad
\delta_H \;=\; \sum_{t=0}^{H-1} A^{H-1-t}\, e_t .
\end{equation}
Unlike the residual, the drift accumulates every per-step error, amplified by the passive dynamics.

\begin{proposition}[Goal-distance bound]
\label{prop:goal-bound}
Let $\rho = \lVert A \rVert_2$. Under Assumption~\ref{assump:identifiability}, with the oracle IDM, every candidate $a_{0:H-1}$ satisfies
\begin{equation}
\label{eq:goal-bound}
\lVert z_H - z_g \rVert \;\leq\; \sqrt{c_g} \;+\; s_{\max}(B) \sum_{t=0}^{H-1} \rho^{\,H-1-t}\, \big\lVert B^{+} e_t \big\rVert \;+\; \sum_{t=0}^{H-1} \rho^{\,H-1-t}\, \big\lVert e_t^{\perp} \big\rVert .
\end{equation}
Moreover, for every weight $w > 0$,
\begin{equation}
\label{eq:goal-bound-any-w}
\lVert z_H - z_g \rVert \;\leq\; \sqrt{\Big( 1 + \frac{\nu}{w} \Big) \big( c_g + w\, c_a \big)} \;+\; \sum_{t=0}^{H-1} \rho^{\,H-1-t}\, \big\lVert e_t^{\perp} \big\rVert,
\end{equation}
where $\nu = s_{\max}(B)^2 H\, \Gamma_H$ and $\Gamma_H = \sum_{k=0}^{H-1} \rho^{\,2k}$. 
\end{proposition}

\begin{proof}
\emph{Eq.~(\ref{eq:goal-bound}).}
Since $z_H = \hat{z}_H - \delta_H$,
\begin{align*}
\lVert z_H - z_g \rVert
&\;\leq\; \lVert \hat{z}_H - z_g \rVert + \lVert \delta_H \rVert \\
&\;\leq\; \sqrt{c_g} \;+\; \sum_{t=0}^{H-1} \rho^{\,H-1-t} \lVert e_t \rVert \\
&\;\leq\; \sqrt{c_g} \;+\; \sum_{t=0}^{H-1} \rho^{\,H-1-t} \Big( \big\lVert e_t^{\sparallel} \big\rVert + \big\lVert e_t^{\perp} \big\rVert \Big) \\
&\;\leq\; \sqrt{c_g} \;+\; s_{\max}(B) \sum_{t=0}^{H-1} \rho^{\,H-1-t} \big\lVert B^{+} e_t \big\rVert \;+\; \sum_{t=0}^{H-1} \rho^{\,H-1-t} \big\lVert e_t^{\perp} \big\rVert .
\end{align*}
The first inequality is the triangle inequality; the second follows from Eq.~(\ref{eq:drift}) and $\lVert A^{H-1-t} \rVert \leq \rho^{\,H-1-t}$; the third splits $e_t = e_t^{\sparallel} + e_t^{\perp}$; and the last applies the right inequality of Eq.~(\ref{eq:two-sided}), $\lVert e_t^{\sparallel} \rVert \leq s_{\max}(B) \lVert B^{+} e_t \rVert$.

\smallskip
\emph{Eq.~(\ref{eq:goal-bound-any-w}).}
Starting from Eq.~(\ref{eq:goal-bound}),
\begin{align*}
\lVert z_H - z_g \rVert
&\;\leq\; \sqrt{c_g} \;+\; s_{\max}(B) \sum_{t=0}^{H-1} \rho^{\,H-1-t} \big\lVert B^{+} e_t \big\rVert \;+\; \sum_{t=0}^{H-1} \rho^{\,H-1-t} \big\lVert e_t^{\perp} \big\rVert \\
&\;\leq\; \sqrt{c_g} \;+\; \sqrt{\nu\, c_a} \;+\; \sum_{t=0}^{H-1} \rho^{\,H-1-t} \big\lVert e_t^{\perp} \big\rVert \\
&\;\leq\; \sqrt{\Big( 1 + \frac{\nu}{w} \Big) \big( c_g + w\, c_a \big)} \;+\; \sum_{t=0}^{H-1} \rho^{\,H-1-t} \big\lVert e_t^{\perp} \big\rVert .
\end{align*}
The second inequality applies the Cauchy--Schwarz inequality with Eq.~(\ref{eq:ca-oracle}):
$\sum_{t} \rho^{\,H-1-t} \lVert B^{+} e_t \rVert \leq \big( \sum_{t} \rho^{\,2(H-1-t)} \big)^{1/2} \big( \sum_{t} \lVert B^{+} e_t \rVert^2 \big)^{1/2} = \sqrt{\Gamma_H}\, \sqrt{H c_a}$, and $s_{\max}(B) \sqrt{\Gamma_H H c_a} = \sqrt{\nu\, c_a}$.
The third applies Young's inequality $2xy \leq \mu x^2 + y^2/\mu$, which gives $(x + y)^2 \leq (1 + \mu)\, x^2 + (1 + 1/\mu)\, y^2$ for any $\mu > 0$. With $x = \sqrt{c_g}$, $y = \sqrt{\nu\, c_a}$, and $\mu = \nu / w$, the right-hand side is $(1 + \mu)\, c_g + (1 + 1/\mu)\, \nu\, c_a = (1 + \mu)(c_g + w\, c_a)$, since $(1 + 1/\mu)\, \nu = \nu + w = (1 + \mu)\, w$.
\end{proof}

\noindent\textbf{Interpretation.}
A candidate whose predicted trajectory drifts toward the goal through world-model error can attain a small $c_g$ while its true terminal state remains far from the goal, and the goal cost alone cannot detect this. Proposition~\ref{prop:goal-bound} shows that the augmented cost closes this gap for the action-explainable part of the drift: for any weight $w_a > 0$, the true goal distance is bounded by the square root of the augmented cost $c_g + w_a c_a$ of Eq.~(\ref{eq:augmented-cost}), scaled by $\sqrt{1 + \nu / w_a}$, plus the deviations that no action can produce. Since this factor depends only on $w_a$, $A$, $B$, and $H$, it is common to all candidates of a CEM population, so ranking candidates by the augmented cost ranks them by the action-explainable part of this upper bound; the remaining term, which neither cost observes, is discussed below. Minimizing the augmented cost therefore minimizes the action-explainable part of a pessimistic estimate of the true goal distance, analogous to the uncertainty-penalized reward of MOPO~\citep{yu2020mopo}.

\noindent\textbf{Relation to the adaptive weight.}
Eq.~(\ref{eq:goal-bound-any-w}) holds for every $w > 0$, so the adaptive weight of Eq.~(\ref{eq:idm-weight}) does not affect the validity of the bound; $w_a$ only sets how tight it is. The choice of $w_a$ itself is governed by the relative spread of the two costs across candidates (Sec.~\ref{sec:planning} and Sec.~\ref{sec:supp_weight}).

\noindent\textbf{The component outside $\mathrm{range}(B)$.}
The last term of Eq.~(\ref{eq:goal-bound}) is not controlled by $c_a$: by Lemma~\ref{lem:pinv}(c), the ideal IDM discards $e_t^{\perp}$. We keep this term rather than assuming it away.
The action consistency cost thus bounds only the \emph{action-explainable} part of the drift, that is, deviations that could have been caused by some action: two candidates whose predictions differ only in $e_t^{\perp}$ receive the same $c_a$, and the augmented cost orders them by $c_g$ alone, as in the baseline.

\subsection{Learned inverse dynamics model and stochastic dynamics}
\label{sec:supp_theory_learned}

\noindent\textbf{Learned IDM.}
In practice we use a learned verifier $G_\phi = G^{*} + \eta$, where $\eta(z, z')$ is its approximation error. Writing $\eta_t = \eta(\hat{z}_t, \hat{z}_{t+1})$, the per-step residual becomes $\lVert a_t - G_\phi(\hat{z}_t, \hat{z}_{t+1}) \rVert^2 = \lVert B^{+} e_t + \eta_t \rVert^2$, which expands as
\begin{equation}
\label{eq:learned-residual}
\big\lVert B^{+} e_t + \eta_t \big\rVert^2 \;=\; \underbrace{\big\lVert B^{+} e_t \big\rVert^2}_{\text{world-model error}} \;+\; \underbrace{\lVert \eta_t \rVert^2}_{\text{IDM floor}} \;+\; \underbrace{2\,\big\langle B^{+} e_t,\, \eta_t \big\rangle}_{\text{cross term}} .
\end{equation}

The first term ranks candidates by the world model's error as analyzed above. The second term is an approximation floor of the IDM, which enters the ranking only through its variation across candidates~(Sec.~\ref{sec:supp_weight}). Within a population, every candidate is sampled from the same Gaussian around the current mean sequence and rolled out from the same $z_0$, so the transitions $(\hat{z}_t, \hat{z}_{t+1})$ the IDM evaluates lie in a common neighborhood, over which $\eta$, a continuous function of these transitions, varies smoothly; the floor therefore acts as a near-common offset as long as its spread across candidates is small relative to that of the world-model error term. The cross term is the only way the IDM could mask a world-model error, and it would require $\eta_t$ to be systematically anti-aligned with $B^{+} e_t$ across candidates. $G_\phi$ is trained on real transitions, independently of $F_\theta$, to predict actions rather than states, and no loss couples the two models; nothing in either objective induces such an alignment. We note that both conditions, a small spread of the floor and the absence of such an alignment, are assumptions rather than guarantees; Sec.~\ref{sec:exp_idm_degrad} examines how much IDM error the gain tolerates in practice.

\noindent\textbf{Identifiability, not determinism.}
Assumption~\ref{assump:identifiability} constrains how actions register in the latent space, not whether transitions are deterministic. With additive noise, $z_{t+1} = A z_t + B a_t + \xi_t$ with $\xi_t$ independent of $a_t$, the least-squares action of a real transition is $B^{+}(z_{t+1} - A z_t) = a_t + B^{+} \xi_t$. It deviates from $a_t$ by $B^{+} \xi_t$, whose distribution does not depend on the executed action, so the action remains identifiable up to a noise level set by $\xi_t$ and $s_{\min}(B)$. Proposition~\ref{prop:residual} continues to hold with $f$ taken as the noise-free part of the dynamics, and Proposition~\ref{prop:goal-bound} holds with an additional term $\sum_{t=0}^{H-1} \rho^{\,H-1-t} \lVert \xi_t \rVert$ on its right-hand side, which does not depend on the candidate and thus leaves the ranking unchanged. Consistent with this, the action consistency cost still improves planning on goal-conditioned visual navigation with NWM (Sec.~\ref{sec:supp_nwm}), whose world model and IDM are trained on real-world trajectories with inherently stochastic transitions rather than on noise-free ones.

\section{Implementation Details of IDM}
\label{sec:supp_idm}

\begin{figure}[h]
    \centering
    \includegraphics[width=0.55\columnwidth]{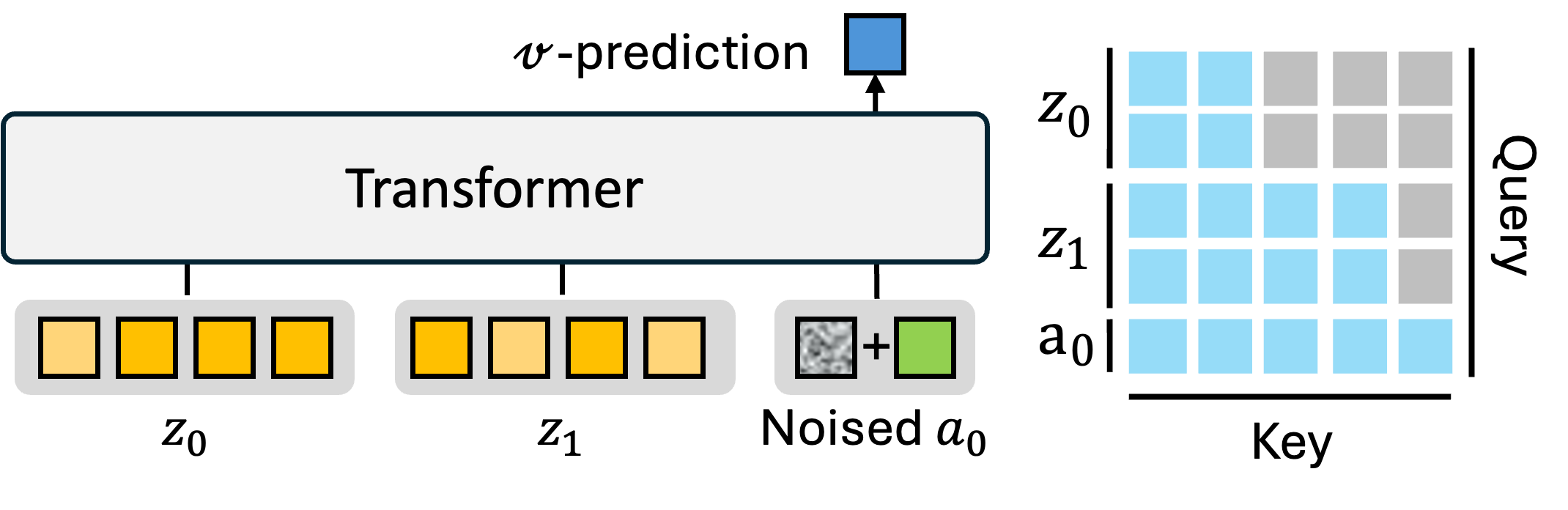}
    \vspace{-3mm}
    \caption{\textbf{Inverse Dynamics Model (IDM) architecture.} (\textit{Left}): the prefix--suffix transformer that predicts the action between the latents of two consecutive frames $z_t$ and $z_{t+1}$. (\textit{Right}): the corresponding attention mask.}
    \label{fig:qual_idm}
\end{figure}
\textbf{Architecture.}
In Fig.~\ref{fig:qual_idm}, we illustrate the IDM, including its model architecture and attention mask.
The IDM is a 4-layer transformer with 3 heads and a width of 192.
The latents $z_t$ and $z_{t+1}$ form the prefix, and a single suffix token carries the noisy action $x_\tau$.
The prefix attends only to itself, while the suffix attends to both prefix tokens; the suffix output is linearly mapped to the velocity in the action space.
We use a flow matching objective~\citep{lipman2022flow, liu2022rectified} with timesteps drawn from a $\mathrm{Beta}(1.5, 1.0)$ schedule.

\textbf{Training data.}
We train a separate IDM for each task, using that task's offline dataset and the frozen encoder latents of the corresponding world model.
Each example is a transition $(z_t, z_{t+1}, a_t)$, where $z_t, z_{t+1}$ are the encoder latents of two consecutive frames and $a_t$ is the action between them; transitions are formed only within episode boundaries.
Actions are standardized to zero mean and unit variance using per-dimension training-set statistics, and de-normalized at inference.
We hold out $10\%$ of the transitions as a validation split.

\textbf{Training and inference.}
IDM is trained for 200K steps with batch size 256. We use AdamW~\citep{loshchilov2017decoupled} with peak learning rate $10^{-4}$ ($\beta_1=0.9$, $\beta_2=0.999$, weight decay $10^{-4}$), with linear warmup followed by cosine decay.
During inference, we integrate the learned velocity field with a single Euler step for PushT, Reacher, OGBench-Cube, Rope manipulation, and Granular manipulation.

\section{World model baselines}
\subsection{Model descriptions}
We conduct experiments with three JEPA-style latent predictors that differ in how
their latent spaces are acquired. They share the
same structure---an encoder maps a frame observation to a latent, and a predictor
predicts the next-frame latent from the current latent and action---and differ
mainly in how the latent space is obtained and how representation collapse is
prevented.

\textbf{Le-WM}~\citep{maes2026leworldmodel} learns the encoder and predictor
jointly from pixels, with a ViT encoder and a transformer predictor that receives
the action through adaptive layer normalization. It prevents collapse with a
single regularizer, SIGReg~\citep{balestriero2025lejepa}, which encourages the
latent embeddings to follow an isotropic Gaussian by applying a normality test
along random one-dimensional projections, in place of the multiple regularizers
used by prior end-to-end approaches.

\textbf{PLDM}~\citep{sobal2026learning} also trains the encoder and predictor
jointly from pixels, optionally using an ensemble of predictors. It prevents
collapse with a VICReg-family objective that combines variance and covariance
regularization with temporal regularization and an inverse dynamics modeling
(IDM) term, so its training objective is composed of several regularizers.

\textbf{DINO-WM}~\citep{zhou2024dino} instead learns only the predictor over
frozen features from a pretrained DINOv2 encoder, sidestepping collapse through
the fixed pretrained representation; we disable its non-visual (e.g.,
proprioceptive) inputs so that all methods observe only pixels unless otherwise
noted.

\subsection{Training configuration} 

\noindent\textbf{PLDM \& Le-WM.~}
For PushT, Reacher, and OGBench-Cube, we use the official pretrained Le-WM checkpoints released by the authors~\citep{maes2026leworldmodel}, and train PLDM with the codebase and hyperparameters reported in the Le-WM paper.
Since neither the original Le-WM nor PLDM paper evaluates on the Rope and Granular tasks, we add these tasks ourselves and train both models on the dataset released by DINO-WM~\citep{zhou2024dino}, following the same training configuration as DINO-WM.

\noindent \textbf{DINO-WM.~}
For OGBench-Cube, Reacher, and PushT, we train DINO-WM for as many epochs as needed to reach the performance reported in \citet{maes2026leworldmodel}.
For the Rope and Granular tasks, we follow the training configuration and official codebase of the original DINO-WM paper~\citep{zhou2024dino} without modification.

The dataset details and training configurations of all three world models are summarized in Tab.~\ref{tab:wm_hparams}.

\begin{table}[t]
\centering
\caption{\textbf{Dataset details and world model training configurations.} \emph{Size} is the number of trajectories and \emph{\# Frames} the trajectory length. History length, frame skip, and train epochs are for training the world model; history length and frame skip are shared across the three world models. For Le-WM on PushT, Reacher, and OGBench-Cube, we use the official pretrained checkpoints, and the listed epochs are those used to train them.}
\label{tab:wm_hparams}
\small
\begin{tabular}{lccccccc}
\toprule
 & & & & & \multicolumn{3}{c}{Train epoch} \\
\cmidrule(lr){6-8}
Dataset & Size & \# Frames & History len & Frame skip & Le-WM & PLDM & DINO-WM \\
\midrule
PushT        & 18685 & 125 (avg.) & 3 & 5 & 10  & 10  & 50  \\
Reacher      & 10000 & 201        & 3 & 5 & 10  & 10  & 50  \\
OGBench-Cube & 10000 & 201        & 3 & 5 & 10  & 10  & 10  \\
Rope         & 1000  & 20         & 1 & 1 & 100 & 100 & 100 \\
Granular     & 1000  & 20         & 1 & 1 & 100 & 100 & 100 \\
\bottomrule
\end{tabular}
\end{table}

\section{Experimental settings}
\label{sec:supp_exp_settings}
\subsection{Environments and tasks}
We evaluate our method across five continuous-control benchmarks that span a spectrum of difficulty, ranging from rigid and deformable object manipulation to articulated control; goal-conditioned visual navigation with NWM is described in Appendix~\ref{sec:supp_nwm}. Each task poses distinct challenges for world-model-based planning, from reasoning about contact dynamics to coordinating deformable objects. The specifics of each environment and its associated dataset are described below.

\textbf{PushT} is a contact-rich 2D manipulation task in which a controllable agent, depicted as a blue circle, must push a green T-shaped block until it aligns with a prescribed target configuration marked in gray. The agent interacts with the block solely through pushing, and successful completion demands an accurate model of the contact dynamics arising between the agent and the object—a property that makes this task a demanding benchmark for object manipulation. We adopt the dataset and protocol of Le-WM~\citep{maes2026leworldmodel}, comprising 18,685 expert episodes whose trajectories average 125 steps in length.

\textbf{Reacher} is a continuous-control task drawn from the DeepMind Control Suite~\citep{tassa2018deepmind}, in which a two-jointed robotic arm must be actuated so that it reaches a designated target within a 2D plane.
Following DINO-WM~\citep{zhou2024dino}, we adopt the more stringent variant in which success requires the full arm configuration—rather than merely the end-effector—to align precisely with the goal pose.
The training dataset consists of 10,000 episodes of 200 steps each, gathered using a Soft Actor-Critic policy. 

\textbf{OGBench-Cube} is a continuous 3D robotic manipulation task originally proposed by OGBench~\citep{park2025ogbench}, in which a robotic arm equipped with an end-effector must grasp a cube and relocate it to a specified target location. Unlike the planar tasks above, this environment requires reasoning over full 3D spatial geometry as well as the coordinated sequence of grasping and placing. We restrict our attention to the single-cube variant. Trajectories are produced using the data-collection heuristic supplied by the benchmark library, yielding 10,000 episodes of 200 steps each.

\textbf{Rope Manipulation} is a deformable-object manipulation task originally introduced by AdaptiGraph~\citep{zhang2024adaptigraph}, in which an XArm interacts with a soft rope placed on a tabletop, simulated in Nvidia Flex. The objective is to manipulate the rope from an arbitrary initial configuration toward a goal configuration specified at test time. Unlike rigid-object tasks, this setting probes whether the world model can faithfully capture the deformable dynamics of the rope together with the contact behavior induced by the robot's pushing actions. We use the dataset released by DINO-WM~\citep{zhou2024dino}, which consists of 1,000 trajectories, each spanning 20 timesteps of random actions initiated from random starting positions, while testing uses goal configurations drawn from a range of initial states with random perturbations in orientation and spatial placement.

\textbf{Granular Manipulation} is a multi-particle manipulation task built on the same simulation backend as the rope task, in which the robot must reshape a collection of roughly one hundred granular particles into a desired target arrangement. Because the particles are disjoint and respond independently to direct contact, this environment stresses the model's capacity to predict the collective dynamics of many loosely coupled objects under physical interaction. We use the dataset released by DINO-WM~\citep{zhou2024dino}, which consists of 1,000 trajectories of 20 timesteps of random actions, all launched from a common initial configuration, while testing is performed on specific target shapes from diverse starting states with random variation in particle distribution, spacing, and orientation.

\subsection{{Details of evaluation protocol}}
For PushT, OGBench-Cube, and Reacher, we follow the goal-conditioned control evaluation protocol of
the original Le-WM paper~\citep{maes2026leworldmodel} without modification for all three world models. Each
evaluation episode is defined by two quantities: the evaluation budget, i.e., the
maximum number of actions the agent may execute in the environment, and the goal
distance, i.e., how many timesteps in the future the goal state is sampled
relative to the initial state. The initial state is obtained by randomly sampling
a state from a trajectory in the offline dataset, and the goal state is the state
occurring a fixed number of timesteps later in the same trajectory, which ensures
that the goal is reachable and consistent with the dataset dynamics. In all three tasks,
the evaluation budget is $50$ steps and the goal is
sampled $25$ timesteps in the future.

For the Rope and Granular manipulation tasks, we likewise follow the evaluation protocol of the original
DINO-WM paper~\citep{zhou2024dino} without modification for all three world models. In both tasks,
the agent must push a deformable object into a
target configuration, and performance is measured by the Chamfer Distance (CD)
between the achieved and goal states rather than a binary success rate. Due to
the high environment stepping cost of these two environments, each is evaluated
on $10$ instances. For granular manipulation, each instance samples a random
initial configuration from the validation set, with the goal of pushing the
material into a square shape at a randomly selected location and scale.

\subsection{Planning solver}
\label{supp:planning-hparams}
For each task, we apply the same planning configuration to all three world models.
For Cube, Reacher, and PushT, we follow the planning configuration of the original paper~\citep{maes2026leworldmodel}. We use a CEM solver that samples 300 action sequences per iteration and runs 30 optimization steps. In each step, the 30 candidates with the lowest cost are selected as elites to update the sampling distribution.

For the Rope and Granular tasks, we use a CEM solver that samples 200 action sequences per iteration for the Rope task and 100 action sequences for the Granular task, with both tasks running 10 optimization steps. In each step, the 30 candidates with the lowest cost are selected as elites to update the sampling distribution. Because the original DINO-WM paper~\citep{zhou2024dino} did not release official planning settings for the Rope and Granular tasks, we selected the settings with which DINO-WM most closely matched the performance reported in that paper.

For the consistency weight, we use $\lambda = 0.07$ for Le-WM, $\lambda = 0.1$ for PLDM, and $\lambda = 0.5$ for DINO-WM on Cube, Reacher, and PushT, and $\lambda = 0.5$ for all three world models on Rope and Granular.
The NWM setting uses $\lambda = 0.08$, and for the real-robot experiments we use $\lambda = 1$ directly, without hyperparameter search.
Because \modelname{} is insensitive to the exact value of $\lambda$ (Table~\ref{tab:robustness}, right; Sec.~\ref{subsec:lambda_exp} of the main), these values need not be finely tuned.

\subsection{Decoders of Le-WM and PLDM, for visualization only}

We train a lightweight transformer decoder that maps a frozen world-model
CLS embedding back to pixels, used \emph{for visualization only}. The CLS
embedding serves as a single memory token attended to by $196$ learnable
patch queries through a $4$-layer, $8$-head Transformer decoder ($H{=}512$),
with each output token linearly mapped to a $16{\times}16$ RGB patch. Since
the decoder sees only one vector, it cannot exploit encoder spatial
features. We freeze each world model's encoder and projector and train the
decoder with MSE on the corresponding expert dataset. Architecture, hyperparameters, and data are shared across
Le-WM and PLDM, so qualitative differences reflect the representations,
not the decoder.

\section{Planning with a video generation-based world model}
\label{sec:supp_nwm}

To test whether \modelname{} extends beyond JEPA-style latent predictors, we apply it to NWM trained with CompACT~\citep{bar2025navigation,kim2026planning}, a world model based on video generation, on goal-conditioned visual navigation on RECON~\citep{shah2021rapid}.
As in the main experiments, we keep the world model and the CEM planner fixed and vary only the planning cost.

\noindent\textbf{World model.}
NWM~\citep{bar2025navigation} uses a Conditional Diffusion Transformer to predict future observations from past observations and navigation actions, and we replace its SD-VAE tokenizer with CompACT-16~\citep{kim2026planning}, which encodes each observation into only $16$ discrete tokens to keep decision-time prediction tractable.
We use the checkpoint released by the authors, trained with the 16-token version of the CompACT tokenizer, instead of reproducing the training.

\noindent\textbf{Task and evaluation.}
Goal-conditioned visual navigation frames planning as an image-goal problem: given a current observation and a target image, the agent searches over action sequences by rolling out predicted future observations until it finds a path that reaches the goal.
We evaluate on the RECON dataset~\citep{shah2021rapid}, a large-scale collection of more than 5000 trajectories gathered autonomously in diverse real-world settings, with substantial visual variation driven by changing seasons and lighting conditions. Of the multiple sensing modalities available, we rely solely on the stereo RGB camera streams.
We adopt the same experimental configuration as the original CompACT paper~\citep{kim2026planning} and report two complementary metrics. Absolute Trajectory Error (ATE) captures the global fidelity of an estimated trajectory by measuring the Euclidean distance between matched points on the predicted and ground-truth paths, while Relative Pose Error (RPE) gauges local consistency by quantifying the discrepancy in the relative transformation between successive poses~\citep{sturm2012evaluating}.

\begin{table}[t]
\centering
\caption{\textbf{Planning performance on goal-conditioned visual navigation with NWM trained with CompACT.} The metrics are Absolute Trajectory Error (ATE) and Relative Pose Error (RPE), and lower is better for both. \emph{Original} and \emph{Ours} are as in Tab.~\ref{tab:main}, and parenthesized values denote the relative change (\%).}
\scalebox{0.9}{
\begin{tabular}{llcc}
\toprule
Model & Planning cost & ATE & RPE (trans.) \\
\midrule
\multirow{2}{*}{NWM w/ CompACT} & Original & 1.3141 & 0.3831 \\
                         & Ours    & \textbf{1.2835 \,(-2.3\%)} & \textbf{0.3773 \,(-1.5\%)} \\
\bottomrule
\end{tabular}
}
\label{tab:nwm}
\end{table}

\noindent\textbf{Planning.}
The world model plans with a distance to the goal between observations~\citep{zhang2018unreasonable} as its goal cost.
Following the planning configuration of the original paper~\citep{bar2025navigation}, we use a CEM solver that samples 80 action sequences per iteration and runs a single optimization step, selecting the 5 lowest-cost candidates as elites.
The IDM integrates its learned velocity field with ten Euler steps.

\begin{figure}[t!]
    \centering
    \includegraphics[width=0.97\columnwidth]{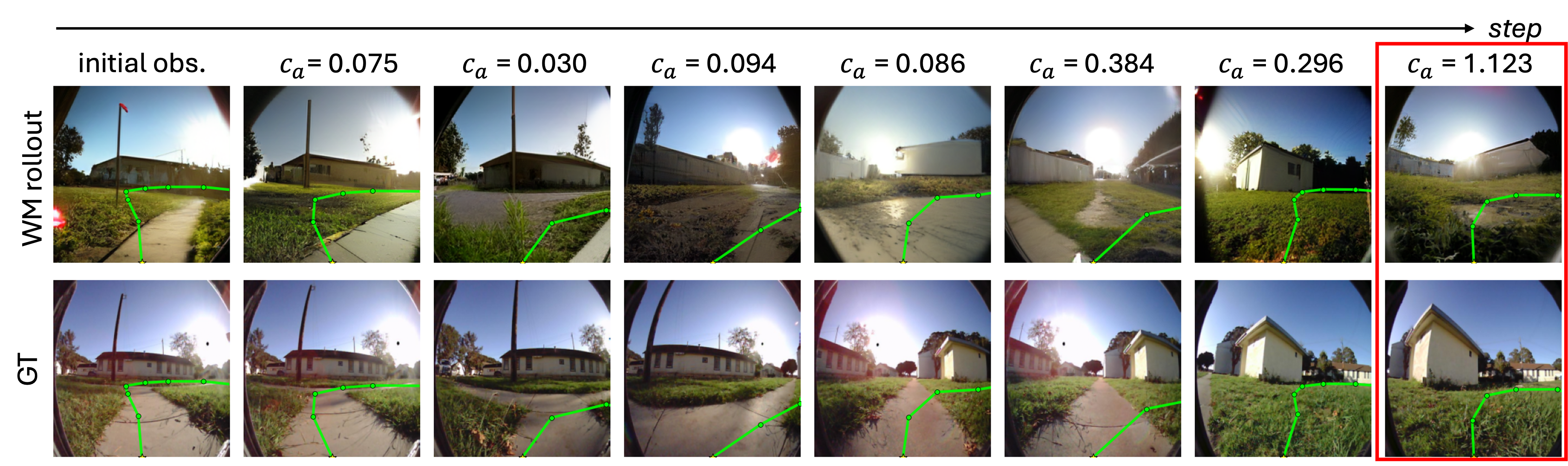}
    \vspace{-3mm}
    \caption{\textbf{Qualitative results on goal-conditioned visual navigation.} Comparison between ground-truth frames and world model rollouts. The action consistency score from \modelname (\textit{top}) and navigation trajectories (\textit{green}) are presented together. Red box highlights the step where the action consistency cost rises; at this step, the predicted next frame fails to follow the action condition.}
    \label{fig:qual_nwm}
\end{figure}

\noindent\textbf{Results.}
Tab.~\ref{tab:nwm} shows that adding the action consistency cost lowers ATE by $2.3\%$ and RPE by $1.5\%$ relative to \emph{Original}.
The gain is smaller than on the manipulation tasks, but it is obtained without any change to the world model, indicating that the action consistency cost applies to world models that generate future observations as well as to those that predict latent features.

\noindent\textbf{Qualitative results.}
Since no environment rollout is available in this setting, Fig.~\ref{fig:qual_nwm} instead compares the ground-truth video against the world model's imagined rollout under the ground-truth actions: starting from the initial observation, we feed the ground-truth action sequence to the world model and compare its predicted frames against the ground-truth video.
For each frame, we overlay the per-step action consistency score.
The score stays low over the early frames, where the imagined rollout closely follows the ground-truth video, and rises toward the end, where the world model's rollout diverges from it.
This shows that the action consistency score reliably identifies the unrealizable transitions, at which the world model fails to realize its conditioning action.

\section{Details of the real robot experiment}
\label{sec:supp_real_robot}
\begin{figure}[t!]
    \centering
    \includegraphics[width=0.9\columnwidth]{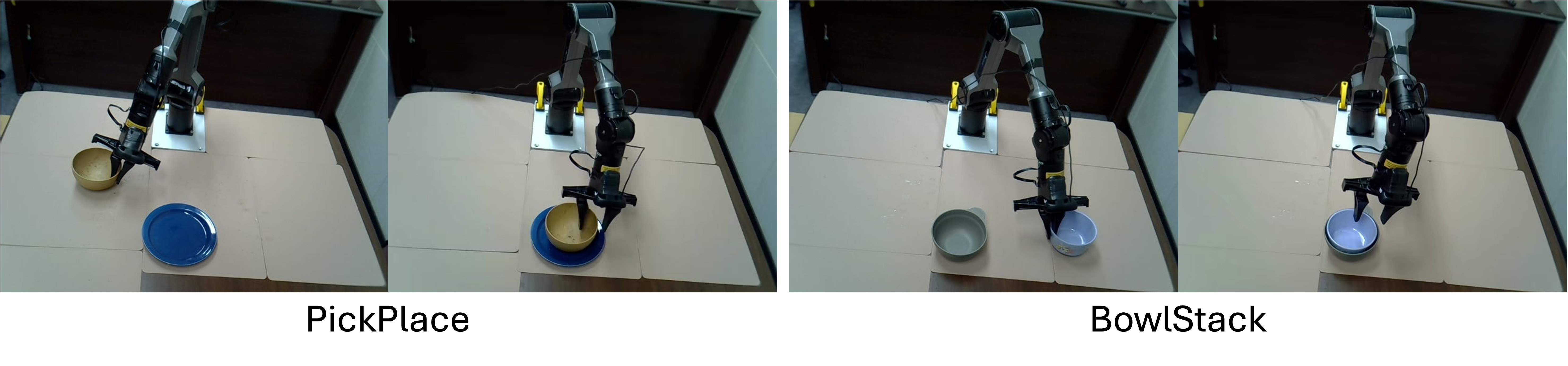}
    \vspace{-3mm}
    \caption{\textbf{Subgoal images of PickPlace and BowlStack.} From left to right: the pick and place goals of PickPlace, then those of BowlStack.}
    \label{fig:robot_goal}
\end{figure}

\noindent\textbf{Robot and data.}
We use a PiPER arm with six joints and a gripper, observed by a single fixed RGB camera; the observation also includes the 7-dimensional joint state, and an action is an absolute joint target of the same dimension.
Demonstrations are collected by teleoperation at 30\,fps in the LeRobot format~\citep{cadene2026lerobot}.
For PickPlace, we collect 200 episodes with the plate at a fixed position and the bowl at 57 distinct initial positions.
For BowlStack, we collect 72 episodes, placing one bowl at a distinct initial position in each episode while keeping the other bowl at a fixed position.
For each task, the world model and the IDM are trained on all of its episodes, and the diffusion policy is trained on a 30-episode subset covering 30 bowl positions for PickPlace and on all 72 episodes for BowlStack.
Each task is specified by two subgoal images, a pick goal and a place goal, as shown in \Figref{fig:robot_goal}.

\begin{table}[t]
\centering
\caption{\textbf{Hyperparameters of Diffusion Policy.}}
\label{tab:dp_hparams}
\scalebox{0.8}{
\begin{tabular}{ll}
\toprule
Name & Value \\
\midrule
Observation horizon & 2 \\
Prediction / execution horizon & 16 / 15 \\
DDIM steps (training / deployment) & 100 / 10 \\
Optimizer & AdamW \\
Learning rate & $10^{-4}$ \\
Batch size & 64 \\
EMA & Yes \\
\bottomrule
\end{tabular}
}
\end{table}

\noindent\textbf{Proposal policy.}
The policy is a 1D-UNet Diffusion Policy~\citep{chi2023diffusionpolicy} with a ResNet-18 image encoder, conditioned on the two most recent camera frames and joint states.
It predicts a 16-step action chunk at 10\,Hz, of which the first 15 steps are executed.
At deployment, each of the $N$ candidates is sampled with a deterministic DDIM sampler from its own initial noise, so the diversity among the candidates comes only from the initial noise.
The hyperparameters are listed in Tab.~\ref{tab:dp_hparams}.

\begin{table}[!htb]
\centering
\caption{\textbf{Dataset details and world model training configurations for the real robot experiment.} Frames are recorded at 30\,fps. History length, frame skip, and train epochs are for training the world model.}
\label{tab:realrobot_wm_hparams}
\small
\begin{tabular}{lccccc}
\toprule
Task & Size & \# Frames & History len & Frame skip & Train epoch \\
\midrule
PickPlace & 200 & 325$\sim$638 & 3 & 45 & 10 \\
BowlStack & 72  & 258$\sim$395 & 3 & 15 & 10 \\
\bottomrule
\end{tabular}
\end{table}

\noindent\textbf{World model and inverse dynamics verifier.}
The world model follows the DINO-WM architecture~\citep{zhou2024dino}, with the frozen DINOv2 encoder replaced by a ResNet encoder trained jointly with the predictor under the temporal straightening regularizer of \citet{wang2026temporal} with weight $0.1$.
The encoder maps each frame to a $16\times16$ grid of 8-dimensional tokens, and the joint state and the action chunk are each embedded to 10 dimensions and concatenated to every visual token.
The predictor is a causal ViT with 6 layers, 16 heads, and MLP width 2048 that takes three past frames and predicts the next latent.
It is trained with AdamW, with a learning rate of $5\times10^{-4}$ for the predictor and $10^{-5}$ for the encoder, and a batch size of 32.
The dataset details and training configurations are listed in Tab.~\ref{tab:realrobot_wm_hparams}.
The IDM is the flow-matching action decoder of Appendix~\ref{sec:supp_idm} with 4 transformer layers and width 192, trained on the frozen latents of this world model; at planning time it recovers the action chunk with a single Euler step.

\clearpage  %
\section{Comparison of CEM and ACID}
We provide the pseudo-code for each test-time optimization method in Algorithm~\ref{alg:cem-baseline} and Algorithm~\ref{alg:cem-ours-color}. Algorithm~\ref{alg:cem-ours-color} is identical in content to the version presented in the main paper, with the components that differ from Algorithm~\ref{alg:cem-baseline} highlighted in a distinct text color.

\begin{algorithm}[H]
\caption{Decision-time Planning with CEM}
\label{alg:cem-baseline}
\begin{algorithmic}[1]
\small
\Require Encoder $E_\theta$, world model $F_\theta$, current observation $o_0$, goal observation $o_g$, planning horizon $H$, \# of samples $N$, \# of elites $K$, \# of iterations $J$
\State Encode states: $z_0 \gets E_\theta(o_0)$,\ \ $z_g \gets E_\theta(o_g)$
\State Initialize sampling distribution $\mathcal{N}(\mu_0, \Sigma_0)$ with $\mu_0 = \mathbf{0}$, $\Sigma_0 = I$
\For{$j = 1$ to $J$}
    \State Sample $N$ action sequences $\{a^{(n)}_{0:H-1}\}_{n=1}^{N} \sim \mathcal{N}(\mu_{j-1}, \Sigma_{j-1})$
    \For{$n = 1$ to $N$}
        \State $\hat{z}_0^{(n)} \gets z_0$
        \For{$t = 0$ to $H-1$}
            \State $\hat{z}_{t+1}^{(n)} \gets F_\theta(\hat{z}_t^{(n)}, a^{(n)}_t)$ \Comment{World model predicted latent state}
        \EndFor
        \State Goal cost $c_g^{(n)} \gets \big\lVert \hat{z}_H^{(n)} - z_g \big\rVert_2^2$
    \EndFor
    \State Select the $K$ sequences with lowest $c_g^{(n)}$ as elites $\mathcal{E}:$ $\mu_j \gets \frac{1}{K} \sum_{n \in \mathcal{E}} a^{(n)}_{0:H-1}$,  $\Sigma_j \gets \mathrm{Var}_{n \in \mathcal{E}}\big(a^{(n)}_{0:H-1}\big)$
\EndFor
\State \Return optimized action sequence $\mu_J$
\end{algorithmic}
\end{algorithm}
\vspace{-4mm}
\begin{algorithm}[H]
\caption{Decision-time Planning with Cycle Action Consistency}
\label{alg:cem-ours-color}
\begin{algorithmic}[1]
\small
\Require Encoder $E_\theta$, world model $F_\theta$, \algnew{inverse dynamics $G_\phi$}, current observation $o_0$, goal observation $o_g$, planning horizon $H$, \# of samples $N$, \# of elites $K$, \# of iterations $J$\algnew{, consistency weight $\lambda$}
\State Encode states: $z_0 \gets E_\theta(o_0)$,\ \ $z_g \gets E_\theta(o_g)$
\State Initialize sampling distribution $\mathcal{N}(\mu_0, \Sigma_0)$ with $\mu_0 = \mathbf{0}$, $\Sigma_0 = I$
\For{$j = 1$ to $J$}
    \State Sample $N$ action sequences $\{a^{(n)}_{0:H-1}\}_{n=1}^{N} \sim \mathcal{N}(\mu_{j-1}, \Sigma_{j-1})$
    \For{$n = 1$ to $N$}
        \State $\hat{z}_0^{(n)} \gets z_0$
        \For{$t = 0$ to $H-1$}
            \State $\hat{z}_{t+1}^{(n)} \gets F_\theta(\hat{z}_t^{(n)}, a^{(n)}_t)$ \Comment{World model predicted latent state}
            \State \algnew{$\hat{a}_t^{(n)} \gets G_\phi(\hat{z}_t^{(n)}, \hat{z}_{t+1}^{(n)})$} \Comment{\algnew{IDM inferred action}}
        \EndFor
        \State Goal cost $c_g^{(n)} \gets \big\lVert \hat{z}_H^{(n)} - z_g \big\rVert_2^2$
        \State \algnew{Action consistency cost $c_a^{(n)} \gets \frac{1}{H}\sum_{t=0}^{H-1}\big\lVert a^{(n)}_t - \hat{a}_t^{(n)}\big\rVert_2^2$}
    \EndFor
    \State \algnew{Adaptive weight $w_a \gets \lambda \cdot \sigma_g / \sigma_a$, where $\sigma_g \gets \mathrm{std}_n\!\big(c_g^{(n)}\big)$,\ \ $\sigma_a \gets \mathrm{std}_n\!\big(c_a^{(n)}\big)$}
    \State \algnew{Augmented planning cost $c^{(n)} \gets c_g^{(n)} + w_a\, c_a^{(n)}$ \ for $n = 1, \ldots, N$}
    \State Select the $K$ sequences with lowest \algnew{$c^{(n)}$} as elites $\mathcal{E}:$ $\mu_j \gets \frac{1}{K} \sum_{n \in \mathcal{E}} a^{(n)}_{0:H-1}$,  $\Sigma_j \gets \mathrm{Var}_{n \in \mathcal{E}}\big(a^{(n)}_{0:H-1}\big)$
\EndFor
\State \Return optimized action sequence $\mu_J$
\end{algorithmic}
\end{algorithm}

\begin{figure}[t!]
    \centering
    \includegraphics[width=0.97\columnwidth]{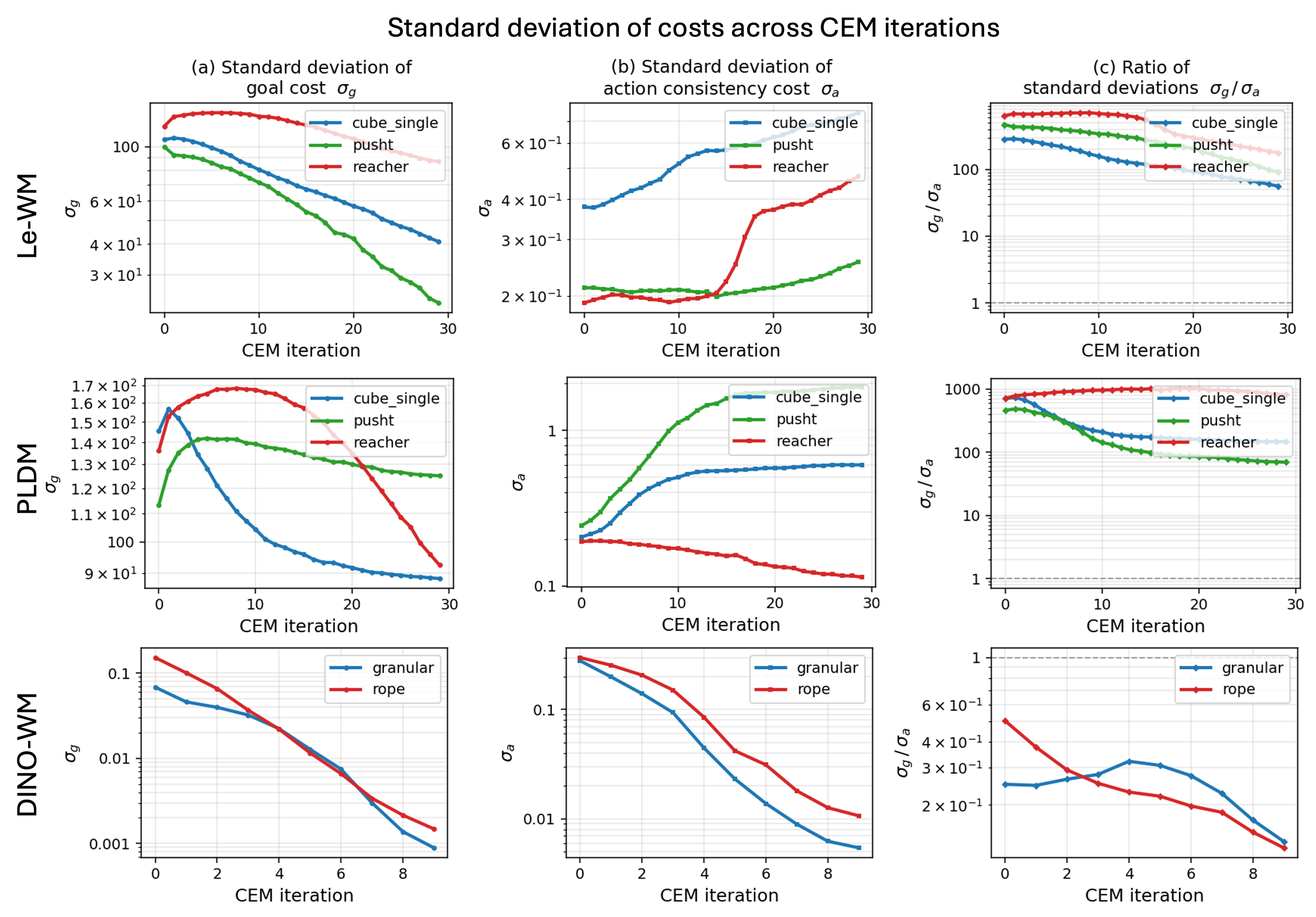}
    \vspace{-3mm}
    \caption{\textbf{Within-pool spread of the goal cost and action consistency cost.}
Rows are world models; columns show the within-pool
standard deviations $\sigma_g$ (a) and $\sigma_a$ (b) and their ratio
$\sigma_g/\sigma_a$ (c) on a log-scaled $y$-axis, with one curve per task. Each
point is the median over all pools at that CEM iteration. The dashed line at
$\sigma_g/\sigma_a=1$ marks equal within-pool spread.}
    \label{fig:cost-std}
\end{figure}

\section{Empirical justification for the scale-invariant adaptive weight}
\label{sec:supp_weight}
In the main section Sec.~\ref{sec:planning} we set the action consistency weight adaptively as
$w_a = \lambda \cdot \sigma_g/\sigma_a$ (Eq.~\ref{eq:idm-weight} of the main),
motivated by the claim that the relative spread of the goal cost $c_g$ and the
action consistency cost $c_a$ varies across (1)~world models, (2)~tasks, and
(3)~CEM iterations. Here we provide the empirical measurements behind that claim
(Fig.~\ref{fig:cost-std}).

\noindent\textbf{Measurement protocol.}
We define a pool as the set of $N$ CEM candidates evaluated for a single
environment at a single (MPC step, CEM iteration). Within each pool we compute
$\sigma_g$, the standard deviation of the goal cost across the $N$ candidates,
and $\sigma_a$, the standard deviation of the action consistency cost. Because
many pools exist at each CEM iteration (one per MPC step per environment), each
plotted point reports the \emph{median} of $\sigma_g$, $\sigma_a$, or their ratio
over all pools at that iteration; we use the median rather than the mean for
robustness to outlier pools. The numbers of candidates $N$ and CEM iterations
differ per world model (see Sec.~\ref{supp:planning-hparams} for the full
settings); the standard deviation is computed within each pool and is therefore
unaffected by these differences in pool count. The $y$-axis is log-scaled in all
panels, and the gray dashed reference at $\sigma_g/\sigma_a = 1$ in panel~(c)
marks the regime where the two costs carry equal within-pool spread.

\noindent\textbf{Why spread, not magnitude, governs reranking.}
CEM ranks candidates by the augmented cost $c_g + w_a c_a$ and keeps the lowest-cost elites, so $c_g$ and $c_a$ jointly determine which candidates become elites.
A constant offset added to every candidate's cost leaves the ranking
unchanged; only \emph{differences} across candidates move the ordering. Hence the
influence of each term on elite selection is set by how widely it varies across
the pool---its standard deviation---not by its absolute level. This is why we
track $\sigma_g$ and $\sigma_a$ rather than the raw cost values.

\noindent\textbf{The relative spread varies along all three axes.}
Figure~\ref{fig:cost-std} confirms the three sources of variation, with one row
per world model (Le-WM, PLDM, DINO-WM) and one curve per task.
(1) \emph{Across world models.} The absolute scale of $\sigma_g$ in column~(a)
differs by orders of magnitude across rows, and the ratio $\sigma_g/\sigma_a$ in
column~(c) sits at a different overall level in each row. Different latent spaces
thus induce inherently different cost distributions.
(2) \emph{Across tasks.} Within a single row---i.e., fixing the world model---the
values of $\sigma_g$ and $\sigma_a$ still differ across tasks, so the two
spreads, and their ratio in column~(c), take task-dependent values.
(3) \emph{Across CEM iterations.} Both $\sigma_g$ and $\sigma_a$ change
continuously over the course of a single optimization run, and the size of the
resulting drift in the ratio is itself world-model-dependent: in Le-WM and PLDM
the two spreads compress at different rates and $\sigma_g/\sigma_a$ in column~(c)
shifts by up to an order of magnitude, whereas in DINO-WM they decay at nearly
matching rates and the ratio drifts more mildly. In every case, however, the
ratio is far from constant across world models and tasks, so a weight calibrated
on one configuration does not transfer to another.

\noindent\textbf{Why the adaptive weight works.}
Because the ratio is far from constant across all three axes, no single fixed $w_a$ can keep the consistency term's contribution $w_a\sigma_a$ at a representative level: it overrides the goal cost where $\sigma_a$ is large and vanishes where $\sigma_a$ is small.
The adaptive weight removes this dependence by recomputing $\sigma_g/\sigma_a$ at every iteration, so the weighted spread $w_a \sigma_a = \lambda \sigma_g$ tracks the goal-cost spread in every pool.
This leaves only $\lambda$ as a free parameter, which encodes how strongly to prioritize the consistency cost relative to the goal cost.
We find empirically that, for each world model, a single $\lambda$ transfers across all tasks of a benchmark suite without per-task retuning (see Sec.~\ref{subsec:lambda_exp} of the main for the supporting experiment).

\begin{table}[t]
\centering
\caption{\textbf{Per-iteration latency of world models and IDM verifiers.}
$t_{\mathrm{WM}}$ is the wall-clock time of the world model rollout and
$t_{\mathrm{IDM}}$ that of the inverse dynamics verifier forward within one CEM
iteration.}
\label{tab:idm_latency}
\scalebox{0.8}{
\begin{tabular}{lccc}
\toprule
World model & $t_{\mathrm{WM}}$ (ms) & $t_{\mathrm{IDM}}$ (ms) & $t_{\mathrm{IDM}}/t_{\mathrm{WM}}$ (\%) \\
\midrule
Le-WM   & 17.54      & 2.54   & 14.5 \\
PLDM    & 17.33      & 2.51   & 14.5 \\
DINO-WM & 2{,}080.50 & 245.25 & 11.8 \\
\bottomrule
\end{tabular}
}
\end{table}
\begin{table}[t]
\centering
\footnotesize
\setlength{\tabcolsep}{6pt}
\caption{\textbf{Ranking candidates with the action consistency cost.} Each of the 19 distractors is the ground-truth action chunk plus Gaussian noise, with a standard deviation equal to that of the actions in the dataset. \emph{Original} ranks the candidates with the goal cost only, while \emph{Ours} uses the augmented cost of \Eqref{eq:augmented-cost}.}
\label{tab:idm_ranking}
\begin{tabular}{l l cc c cc}
\toprule
\multirow{2}{*}{Task} & \multirow{2}{*}{Model} & \multicolumn{3}{c}{Recall@1 (\%)} & \multicolumn{2}{c}{Spearman $\rho$} \\
\cmidrule(lr){3-5} \cmidrule(lr){6-7}
 & & Original & Ours & $\Delta$ & Original & Ours \\
\midrule
\multirow{3}{*}{PushT}
        & PLDM    & 70.3 & 91.0 & $+20.7$ & 0.44 & 0.51 \\
        & DINO-WM & 67.5 & 89.2 & $+21.7$ & 0.42 & 0.48 \\
        & Le-WM   & 68.1 & 89.8 & $+21.7$ & 0.44 & 0.51 \\
\midrule
\multirow{3}{*}{Reacher}
        & PLDM    & 32.1 & 49.9 & $+17.8$ & 0.38 & 0.43 \\
        & DINO-WM & 22.8 & 34.8 & $+12.0$ & 0.33 & 0.37 \\
        & Le-WM   & 34.2 & 54.6 & $+20.4$ & 0.39 & 0.43 \\
\midrule
\multirow{3}{*}{Cube}
        & PLDM    & 75.3 & 95.9 & $+20.6$ & 0.35 & 0.40 \\
        & DINO-WM & 79.5 & 95.9 & $+16.4$ & 0.40 & 0.43 \\
        & Le-WM   & 91.1 & 98.4 & $+7.3$  & 0.35 & 0.39 \\
\bottomrule
\end{tabular}
\end{table}
\section{Per-iteration overhead of inverse dynamics verifier}
\label{sec:supp_latency}
We report the wall-clock cost of the two networks that evaluate the planning cost within one CEM
iteration: the world model rollout and the inverse dynamics verifier.
Each is timed over one CEM iteration with $300$ candidate action sequences and a
planning horizon of $5$.
We report the verifier overhead as the ratio of the two times.
All measurements are taken on a single RTX~6000 Ada GPU.

\begin{figure}[t!]
    \centering
    \includegraphics[width=0.97\columnwidth]{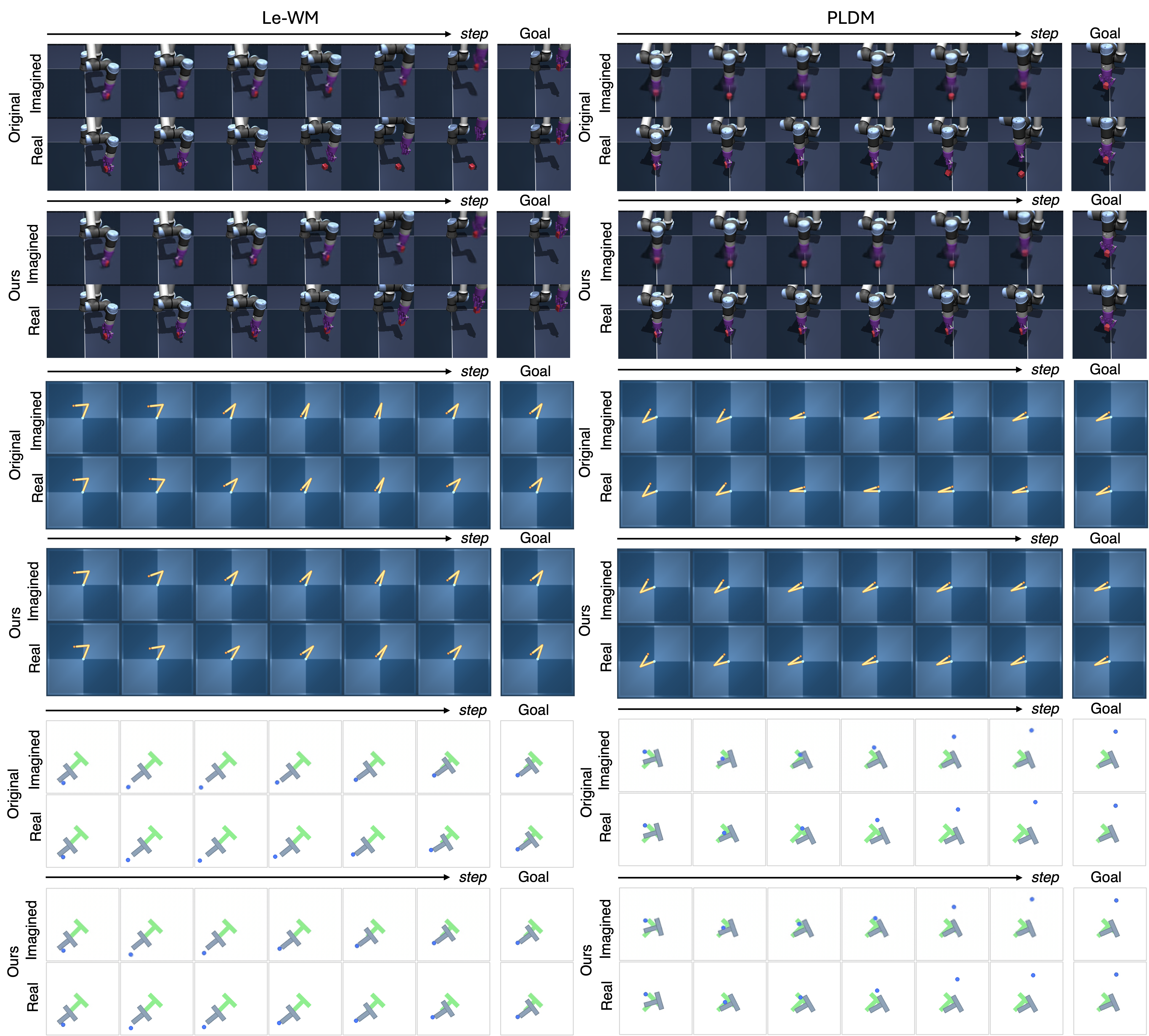}
    \vspace{-3mm}
    \caption{\textbf{Qualitative comparison on Le-WM and PLDM across the OGBench-Cube,
Reacher, and PushT tasks.}
Le-WM is shown in the left column and PLDM in the right column. For each model and
task, we show the real rollout in the environment and the corresponding imagined
rollout from the world model under the planned action sequence, comparing the
baseline with and without \modelname.}
    \label{fig:qual_rollout}
\end{figure}

\section{Ranking candidates with the action consistency cost}
\label{sec:supp_idm_ranking}
Inside the planner, the planning cost scores candidates: the goal cost $c_g$ asks whether a candidate's predicted outcome is close to the goal, while the action consistency cost $c_a$ asks whether that outcome was actually produced by the candidate's actions.
We test whether $c_a$ separates candidates that $c_g$ cannot.

\noindent \textbf{Setup.}
On the validation set of each task, we rank $20$ candidate chunks, the ground-truth chunk and $19$ distractors, using the planner's cost of \Eqref{eq:augmented-cost} with $\lambda = 0.07$.
Each distractor is obtained by adding Gaussian noise to the ground-truth chunk, with a standard deviation equal to that of the actions in the dataset.
We report the recall@1 of the ground-truth chunk in percent and the Spearman $\rho$ between the cost ranking and the action-space distance ranking.
A random ranking yields a recall@1 of $5\%$ and a $\rho$ of $0$.

\noindent \textbf{Results.}
As shown in Tab.~\ref{tab:idm_ranking}, adding $c_a$ raises both recall@1 and $\rho$ for every world model on every task.
Where $c_g$ cannot separate the candidates, $c_a$ restores the ranking by detecting that the action recovered from each prediction disagrees with the candidate that conditioned it.
The IDM thus serves as the verifier the planner requires, consistently across world models and tasks.

\section{Qualitative Results}
In Fig.~\ref{fig:qual_rollout}, we qualitatively compare planning with and without \modelname, for each baseline world model and task.
To enable visual inspection, we train an auxiliary decoder that maps latent states back to pixels, letting us check whether the imagined rollout in latent space agrees with the rollout obtained in the actual environment.
In each figure, the top row shows the pixel-decoded predictions of the world model (the imagined rollout under the planned action sequence), and the bottom row shows the environment rollout obtained by executing that same action sequence.
Without \modelname, the imagined rollout often arises from unrealizable transitions: it reaches a goal-like state in the world model's prediction, but executing the corresponding action sequence drives the environment elsewhere.
With \modelname, the planned action sequence is realizable---the environment rollout closely tracks the imagined rollout, confirming that the predicted goal-reaching can actually be reproduced.

\clearpage

\end{document}